\documentclass[10pt,twocolumn,letterpaper]{article}

\usepackage{cvpr}              

\usepackage{graphicx}
\usepackage{amsmath}
\usepackage{amsfonts}
\usepackage{color}
\usepackage{comment}
\usepackage{enumitem}
\usepackage{amssymb}
\usepackage{booktabs}
\usepackage{appendix}
\usepackage[square, sort&compress, numbers]{natbib}

\usepackage[pagebackref,breaklinks,colorlinks]{hyperref}

\usepackage[capitalize]{cleveref}
\crefname{section}{Sec.}{Secs.}
\Crefname{section}{Section}{Sections}
\Crefname{table}{Table}{Tables}
\crefname{table}{Tab.}{Tabs.}

\newcommand{\bfx}{{\bf x}}

\newcommand{\bfg}{{\bf g}}
\newcommand{\bfi}{{\bf i}}

\newcommand{\bfs}{{\bf s}}

\newcommand{\reals}{{\mathbb R}}

\def\cvprPaperID{2363} 
\def\confName{CVPR}
\def\confYear{2023}

\begin{document}

\title{Programmable Spectral Filter Arrays using Phase Spatial Light Modulators}


\author{Vishwanath Saragadam$^\dagger$, Vijay Rengarajan$^\ddagger$, Ryuichi Tadano$^*$,\\
	 Tuo Zhuang$^*$, Hideki Oyaizu$^*$, Jun Murayama$^*$, Aswin C. Sankaranarayanan$^\ddagger$\\
	$^\dagger$Rice University\\
	$^\ddagger$Carnegie Mellon University\\
	$^*$Sony Group Corporation, Japan\\
	{\tt\small saswin@andrew.cmu.edu}
}

\maketitle

\begin{abstract}
%
%
%
%
%

%
Spatially varying spectral modulation can be implemented using a liquid crystal spatial light modulator (SLM) since it provides an array of liquid crystal cells, each of which can be purposed to act as a programmable spectral filter array.
However, such an optical setup suffers from strong optical aberrations due to the unintended phase modulation, precluding spectral modulation at high spatial resolutions.
In this work, we propose a novel computational approach for the practical implementation of phase SLMs for implementing spatially varying spectral filters.
We provide a careful and systematic analysis of the aberrations arising out of phase SLMs for the purposes of spatially varying spectral modulation.
The analysis naturally leads us to a set of ``good patterns" that minimize the optical aberrations.
We then train a deep network that overcomes any residual aberrations, thereby achieving ideal spectral modulation at high spatial resolution.
We show a number of unique operating points with our prototype including dynamic spectral filtering, material classification, and single- and multi-image hyperspectral imaging.

\end{abstract}

\section{Introduction}
\label{sec:intro}
\begin{figure}
	\centering
	\includegraphics[width=\linewidth]{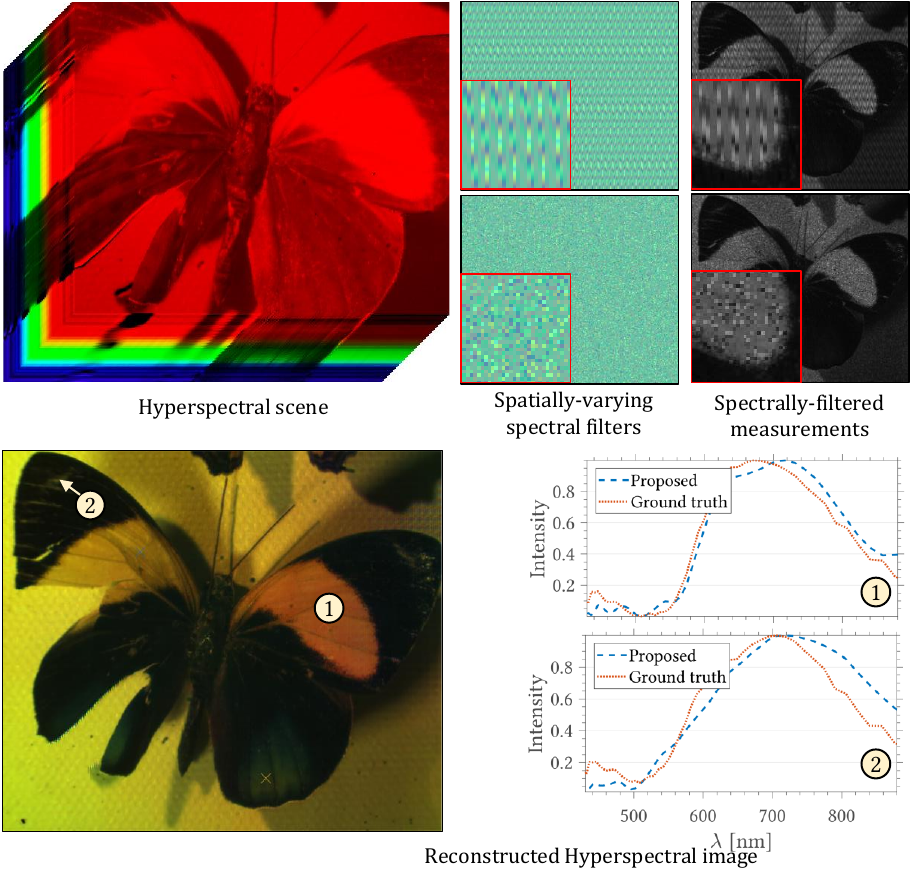}
	\caption{\textbf{Programmable spectral filters.} We propose a practical implementation of spatially varying and programmable spectral filter array.	Our camera can capture intensity images under a wide configuration of spatially varying spectral filters, which plays a critical role in applications that benefit from spectral measurements, namely, hyperspectral imaging, and material classification. Above, we show an example of hyperspectral imaging of a butterfly in visible to near-infrared wavelengths, reconstructed from eight images, each with a different spatial tiling of spectral filters.} \label{fig:teaser}
\end{figure}

Modulating light across its many dimensions is an important enabling capability for many computational imaging and illumination systems.
For example, spatial modulation of light is a critical enabler of projectors.
Similarly, spatio-angular modulation of light, using a micro-lenslet array or a coded aperture finds use in light field imaging.
Hence, it is understandable that many of the advances in computational imaging have come about due to novel mechanisms for manipulating the different dimensions of light.

%

We are particularly interested in techniques that can modulate the spectral content of a scene.
When we look at the space of such techniques, they are largely clustered around two key properties: programmability with devices such as color filter wheels, liquid crystal (LC) tunable filters~\cite{Wu:84}, and dispersion-based systems~\cite{mohan2008agile}, and spatial selectivity by leveraging the concept of assorted pixels~\cite{narasimhan2005enhancing,Lambrechts2014cmos} where a spectral filter is tiled on top of a sensor.
There is, however, a paucity of approaches that are both \textit{programmable} as well as \textit{spatially selective}.
One approach to enable both programmability as well as spatial selectivity in spectral modulation is via the use of an LC-based spatial light modulator (SLM).
An LC SLM is simply a pixellated array of LC cells, where we have (near-)independent control over the phase retardance induced by each pixel; 
this retardance is converted to a spectral modulation  via the use appropriate optical elements.
By optically aligning an image sensor to the SLM,  we can leverage the degrees of freedom provided by the millions of independently-addressable pixels of the SLM to implement a programmable spatially-varying spectral filter.

Liquid crystal SLMs have been used in prior work for enhancing  color gamut \cite{harm2015use}, color and polarization imaging \cite{Tsai:15} as well as hyperspectral imaging \cite{Zhu:13,chen2018computational}.
However, these approaches do not address a number of practical challenges that stem from undesired properties of these SLMs.
%
%
Spectral modulation with LC SLMs is invariably accompanied with a phase modulation to the incident light.
Further, the physics of their manufacturing restricts LC SLMs to only implement  spatially-smooth phase retardances; hence, it is challenging to implement a diverse set of filters over any small region on the SLM.
All of these artifacts are further acerbated when we image over a large spectral range, in part due to the poor  focusing performance of relay lenses.

\paragraph{Contributions.} This paper proposes a system for implementing a  programmable and spatially-varying spectral filter array using LC SLMs, and correcting the aberrations encountered due to non-idealities of the SLM and the associated optics.
%
We study the factors that control undesirable phase modulation and minimize its effects by a careful design of the SLM's retardance curve as well as the patterns displayed on the SLM. 
To further reduce the amount of aberrations, 
we  train a deep network to restore the spectrally-filtered measurements.
We introduce a dataset, acquired using our lab prototype, consisting of images of scenes captured under a variety of spatially-varying filters.

We investigate three  applications to showcase the effectiveness and potential of the proposed approach: hyperspectral imaging  enabled by dynamically changing the bank of filters displayed on the SLM (see Fig.~\ref{fig:teaser}), material classification with task-specific spatially-tiled spectral filters, and the design of arbitrary spectral filters by placing the SLM in the pupil plane of the imaging system.
%
%
%
%

\paragraph{Limitations.} We inherit several limitations due to our use of phase SLMs and LC cells for spectral modulation. 
Spectral filters implemented using individual LC cells are typically restricted to sinusoidal profiles, with a  spectral resolution that is inversely proportional to wavelength.
Further, phase SLMs can only implement spatial patterns that are smooth, which limits the range of filters we can implement.
%
%
%

\section{Prior Work}
\label{sec:prior}
We briefly discuss key results  in spectral modulation, and their use in hyperspectral imaging and  classification. 

\paragraph{Compressive hyperspectral imaging.} 
%
%
%
%
%
%
%
There have been many imaging architectures proposed that use the theory of compressive sensing to speed up HSI sensing; this includes the coded aperture snapshot spectral imager (CASSI)~\cite{gehm2007single, wagadarikar2008single}, spatio-spectral coding~\cite{lin2014dual}, imager with DMD and a single spectrometer~\cite{sun2009compressive}, and a conventional camera equipped with a prism in front of the lens~\cite{baek2017compact}. This paper falls under the broad category of such designs.

\paragraph{Spectral modulation with LC cells.}
Liquid crystal cells and SLMs rely on birefringence where the two orthogonal states of polarization are delayed by varying amount applications in various fields of optics~\cite{beeckman2011liquid,lyot1933optical,ohman1938new}.
%
%
The spectral modulation property of LC cells have been used for compressive sensing of HSIs~\cite{oiknine2019compressive,august2013compressive} but often require several images to recover the HSI.
Spectral modulation with LC cells have also been used for material classification \cite{zhi2019multispectral}; here,  a few carefully-chosen spectral filters are obtained to detect various powders in a scene.
%
%
%
%
The spatially-varying modulation of SLMs has been leveraged in the past for designing color displays~\cite{harm2015use}, and for spectral imaging~\cite{Zhu:13,Tsai:15,chen2018computational,chen2018computational}.
%
%
%
%
%
%
%
However, these  have not modeled or considered any of the aberrations induced in using phase SLMs for implementing spatially-varying filters; we show that the aberrations caused by the phase SLM can severely degrade reconstruction quality. To the best of our knowledge, we are the first to perform a careful analysis of this effect. 

\paragraph{Assorted pixels.}
Assorted pixels \cite{narasimhan2005enhancing} refers to a technique where a grayscale image sensor is augmented by placing an array of filters on top to provide enhanced perception of spectrum, polarization and/or dynamic range.
%
Assorted pixels have been extended to hyperspectral imaging as well, where a narrowband spectral filter array is tiled on top of the sensor \cite{Lambrechts2014cmos}.
%
%
%
%
The main contribution of our work is in the spirit of assorted pixels; but unlike existing work, where the spectral arrays are fixed at fabrication, the proposed optical design permits a \textit{programmable} array of spectral filters which enhances the scope of the technique in many interesting ways.
For the same reasons, we refer to our approach as \textit{programmable assorted pixels (ProAsPix)}.


\section{Programmable Spectral Filter Arrays}
\label{sec:proaspix}
We now discuss the ideas underlying the implementation of spatially-varying spectral filters using a phase SLM.

\subsection{Spectal Filtering with LC cells and SLMs}
\label{sec:lccells}

%
We briefly go over the principle of operation of an LC cell when used to implement a spectral filter.
The reader is referred to \cite{Wu:84}  as well as section~\ref{sec:lc} for a detailed treatment. 
Consider an imaging setup consisting of an LC cell of thickness $d_{LC}$ that is sandwiched between two linear cross polarizers, with their polarization axes oriented at $\pm 45^\circ$ to the LC cell's fast axis.
Suppose that we apply a voltage $v$ across the LC cell
%
which produces a birefringence of $\Delta n(v)$.
Now, \textit{unpolarized} light incident on this setup experiences a spectral filter of the form:
\begin{equation}
m(v, \lambda; d_{LC}) = \frac{1}{2} \left( 1 - \cos\left( 2\pi \frac{\Delta n(v) d_{LC}}{\lambda} \right)\right), \label{eq:lc04}
\end{equation}
where $\lambda$ is the wavelength of light.
The expression in (\ref{eq:lc04}) provides spectral filter observed when an RMS voltage $v$ is applied across the LC cell.
%
%
%
%

\begin{figure}
\centering
\includegraphics[width=0.475\textwidth]{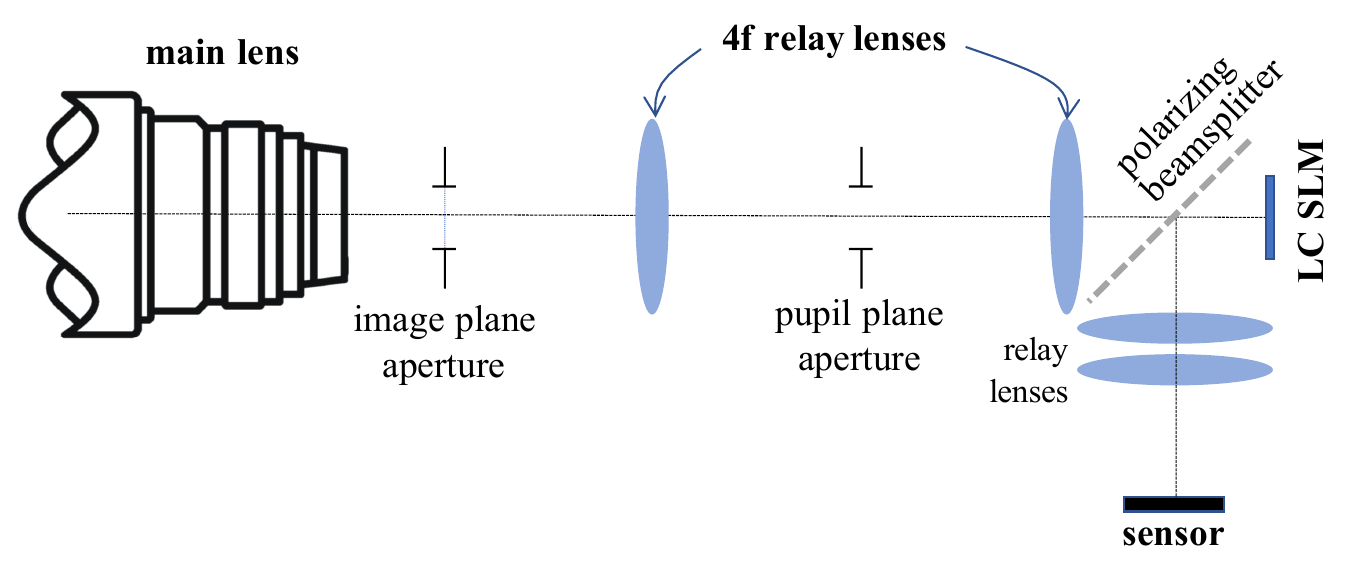}
\caption[]{\textbf{Spatially-varying spectral filtering with SLMs.}   We optically align an LC-based SLM to an image sensor, using optical relays and a polarizing beamsplitter that acts as a pair of cross-polarizers before and after the SLM. This allows us to build a programmable spectral filter array, where a spatially-varying and programmable spectral filter can be realized by displaying different patterns on the SLM.}
\label{fig:schematics}
\end{figure}

%
Our key insight is that an LC SLM is in essence an array of LC cells, each of which acts as a programmable spectral filter.
%
Collocating the SLM with an image sensor, hence, allows us to obtain a device whose spectral response can be changed spatially (see Fig.~\ref{fig:schematics}).
%
%
%
%
%
Suppose we display a spatially-varying voltage pattern $v(x, y)$ on the SLM, then the image $i(x, y)$ observed at the sensor is given as 
\begin{align}
&i(x, y) = \frac{1}{2} \int_\lambda h(x, y, \lambda)  m(v(x, y), \lambda, d_{LC})s(\lambda) d\lambda,\label{eq:proaspix01}\\
&m(v(x, y), \lambda, d_{LC}) = \left(1 - \cos\left( 2\pi \frac{ \Delta n(v(x,y)) d_{LC}}{\lambda} \right) \right)\nonumber
\end{align}
where $s(\lambda)$ is the spectral sensitivity of the image sensor and $h(x, y, \lambda)$ is the unmodulated hyperspectral image formed on the sensor.
Hence, by appropriate choice of the pattern that we display on the SLM, we can implement spatially-varying spectral filters.
The set of filters we can obtain depends on the birefringence of the SLM, the range of input voltage we can provide, and the thickness of the SLM, all of which are device specific.

%
%
%

\begin{figure}
\includegraphics[width=0.475\textwidth]{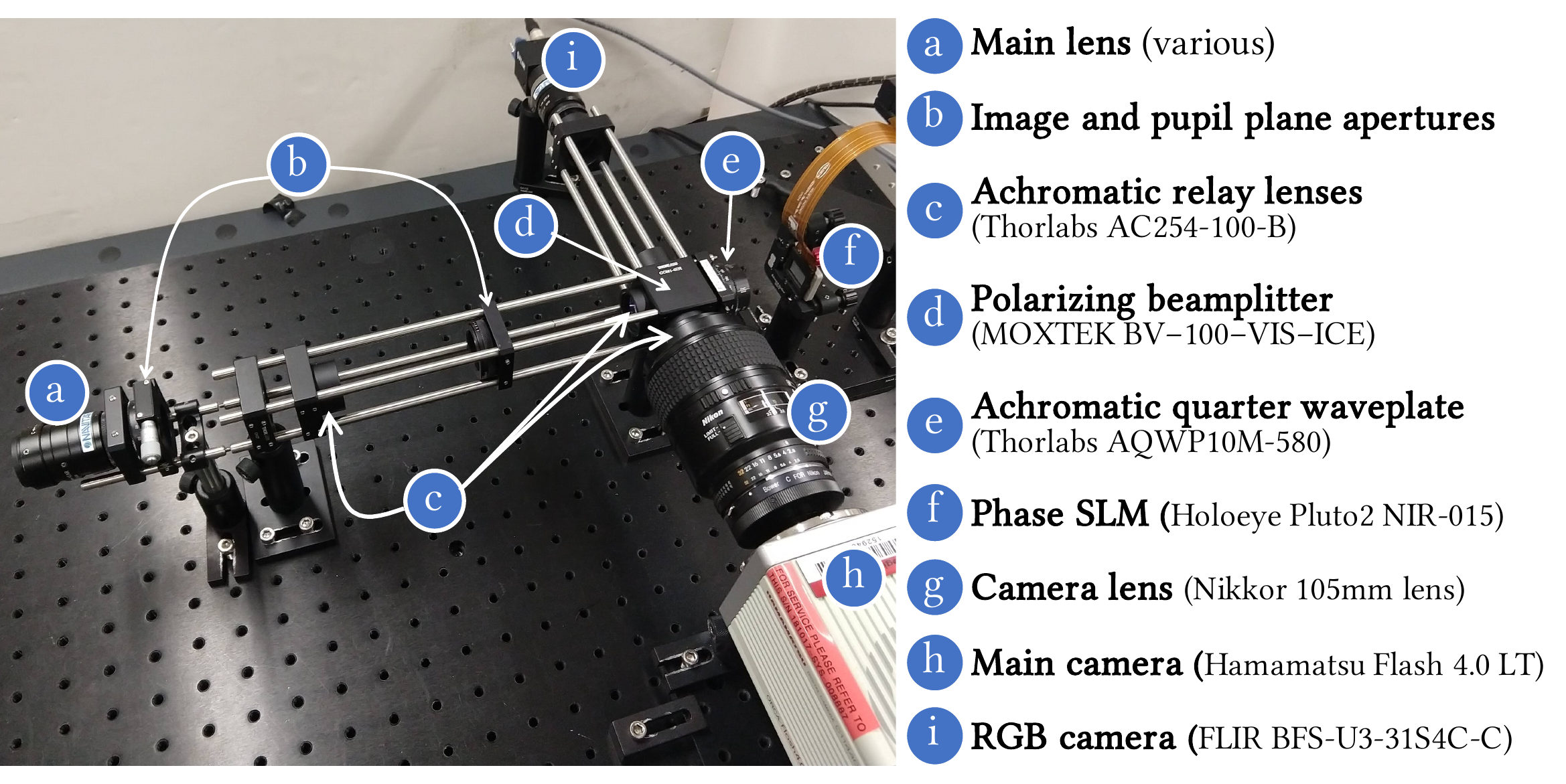}
\caption{\textbf{Lab prototype.}}
\label{fig:proto}
\end{figure}

\paragraph{Hardware prototype.}
%
%
%
Figure \ref{fig:proto} shows the lab prototype, on an optical benchtop, implementing the programmable spectral filter array.
%
%
%
%
%
We use an achromatic quarterwave plate, immediately after the polarizing beamsplitter, with its fast axis aligned at $45^\circ$ to the transmitted polarization state of the beamsplitter.
%
%
 %
%
We have a second image sensor, an RGB unit, placed in the previously unused arm of the beamsplitter, and collocated with the image plane of the system and, hence, the SLM and the other image sensor.
We refer to this as the \textit{guide camera}, and use it for guided filter-based reconstruction techniques that resolve the loss of spatial resolution due to tiling of the spectral filters.
The system is configured to image from 400-1000nm, corresponding to spectral response of the camera.
The lab prototype had a spatial field of $10.4\times8$ sq.mm.
The Fourier (pupil) plan was set to 10mm wide with 100mm relay lens, resulting in an $f/10$ aperture.
A more detailed description, including the approach used for spatial alignment and spectral calibration of the device, are presented in section~\ref{sec:hardware}.

%
%
%

%


\subsection{Linking Phase Gradients to SLM Patterns}
\label{sec:pattern}

\begin{figure}
    \centering
    \includegraphics[width=0.99\linewidth]{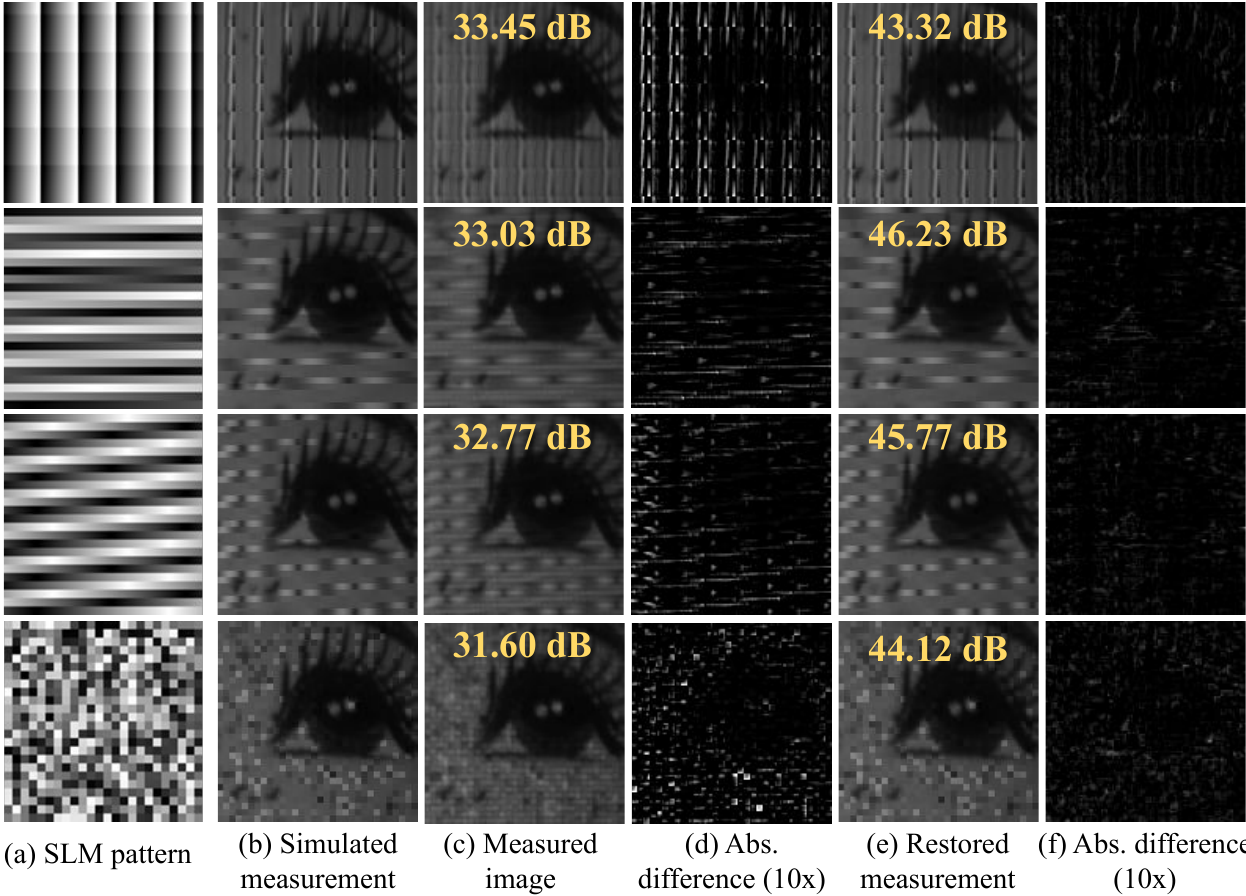}
    \caption{\textbf{Restoration of measured image.} We show $128 \times 128$ patches of a scene from our test data. 
    Each row in the left and right page columns corresponds to some of the different patterns that we employed during capture. The simulated measurement in (a) is the ``ground-truth'' image for each pattern. The measured image in (b) is very much distorted as can be seen by its absolute difference with the simulated measurement in (c). Our restoration network cleans up the distortions and produces the restored measurements shown in (d) that have higher PSNR (the numbers in the inset) as well as less erroneous as can seen in (e).}
    \label{fig:restore_net_outputs}
\end{figure}

%
%
In practice, per-pixel phase modulation with an SLM faces challenges from phase distortions, non-ideal optics, and discontinuities in the SLM pattern.
Since any local phase gradient leads to tilt in light cones, the resultant image is a complex function of the per-pixel phase modulation and the phase gradient itself. 
%
%
%
%
%
%
To understand the effect of these aberrations, we captured an exhaustive set of 256 images for a scene with \textit{constant patterns} on the SLM, thereby resulting in no aberrations, since such patterns do not have phase gradients.
We refer to this set as the \textit{full scan} measurements. 
We can now use this full scan to generate ``simulated measurements" that would represent the ideal measurement with a given SLM pattern (see Fig.~\ref{fig:restore_net_outputs}(b) for each SLM pattern in (a)).
%
%
%
%
%
%
%
%
%
Figure \ref{fig:restore_net_outputs}(c) show the actual measured sensor images for different SLM patterns; we observe that these measurements have a significant mismatch against the  simulated measurements from the full scan data.

The differences between the ideal measurements and the acquired ones can be attributed to strong phase gradients introduced near discontinuities in the pattern displayed on the SLM; for example, consider the random  tiling in row 4 as well as regions of high gradients in row 1 of Fig.~\ref{fig:restore_net_outputs}.
 Consider the spatially-varying phase retardation $\phi(\bfx; \lambda)$ introduced by the SLM; from (\ref{eq:lc04}), this is given as 
\begin{equation}
 \phi(\bfx; \lambda) =  2\pi \frac{\Delta n(v(\bfx)) d_{LC}}{\lambda}.
 \end{equation}
The local distortions to the wavefront can be characterized by the spatial gradients of $\phi$.
%
%
The voltage is related to the pattern as $v(x, y) = \gamma(p(x, y))$
where $p(x, y)$ is the $8$-bit image displayed on the video port and $\gamma(\cdot)$ is the ``gamma curve'' of the SLM, or the mapping of this $8$-bit number to voltage applied at the SLM.
The spatial gradients of $\phi(\bfx; \lambda)$ can be written as
\begin{equation}
\frac{\nabla \phi}{d\bfx} = \frac{2\pi}{\lambda} d_{LC} \frac{\partial \Delta n}{d \gamma} \frac{\partial \gamma}{\partial p} \frac{\partial p}{d \bfx}.
\label{eq:phasegrad01}
\end{equation}
The term $ \frac{\partial \Delta n}{d \gamma}$ is a device-specific property that dictates change in birefringence as a function of cell voltage.
%
%
To simplify the overall design problem, while minimizing the effect of the phase gradient, we choose a gamma curve $\gamma$ for the SLM such that 
\begin{equation}
	\frac{\partial \gamma}{\partial p} = c_0 \left[ d_{LC}   \frac{\partial \Delta n}{d \gamma} \right]^{-1}
	\label{eq:gamma01}
\end{equation}
where $c_0$ is a constant (see section~\ref{sec:gamma} for details). 
%
%
This gives a simplified expression for the phase gradient induced by the SLM:
\begin{equation}
	\frac{\nabla \phi}{d\bfx} = \frac{2\pi c_0}{\lambda} \frac{\nabla p}{d \bfx}.
	\label{eq:phasegrad02}
\end{equation}
%
%
\textit{Hence, the phase distortion is simply a scaled-version of the  gradient of the pattern displayed on the SLM.}

A trivial way to remove distortions is to choose a constant pattern, which lacks spatial diversity and hence is not favorable.
Intuitively, a locally smooth results in smaller spatial gradients, and hence patterns with local ramps (linearly increasing phase) are expected to result in low distortions.
This is visualized in Fig.~\ref{fig:restore_net_outputs} where locally smooth patterns on top achieve higher accuracy than sharply changing ones such as the random pattern in the bottom row.
Invariably, given that phase gradients are unavoidable even for smooth patterns, we resort to computational techniques for handling their adverse affects.

\subsection{Deep Restoration of Sensor Measurements}
\label{sec:restorenet}
\begin{figure}
    \centering
    \includegraphics[width=0.99\linewidth]{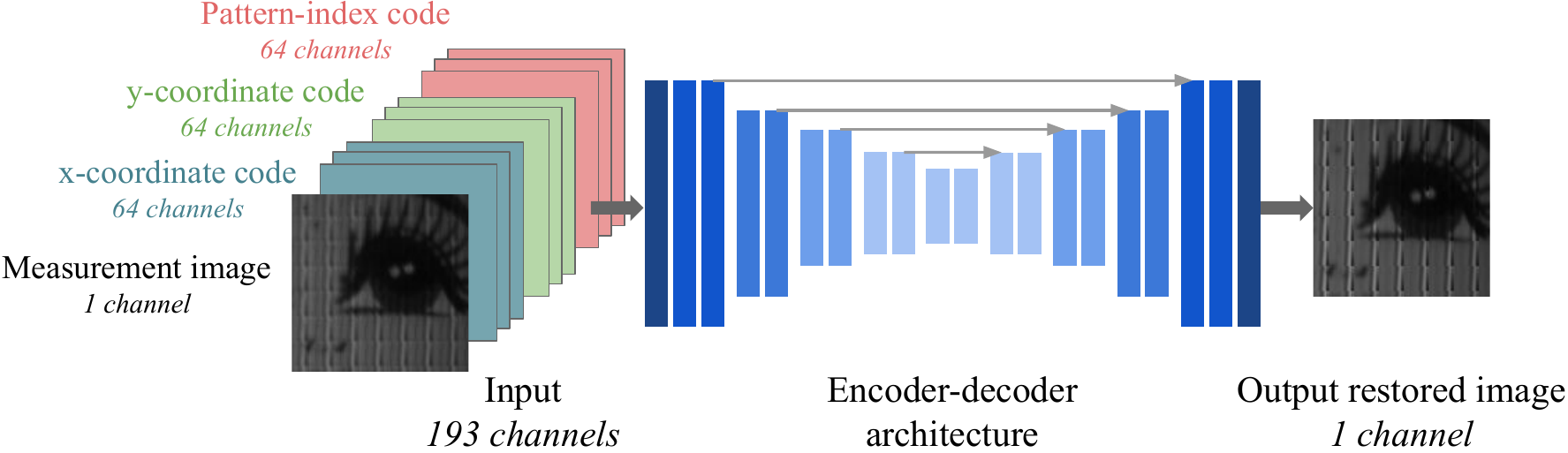}
    \caption{\textbf{Restoration network architecture.} Our pattern-oblivious encoder-decoder architecture rectifies the distortions in the measurement image using positional codes based on coordinates and pattern index.}
    \label{fig:restore_net_arch}
\end{figure}

To fix the aberrations introduced by the non-idealities in the SLM, we devise a learning-based restoration scheme whose goal is to learn the non-linear mapping that takes the distorted measurements and produce clean ``simulated" measurements. 
%
%
%
%
%
%
%
%
%
%
We design a single neural network that takes the measured image as input, for any spatially-varying pattern on the SLM that is also provided as input, and outputs the ``simulated" measurement for that pattern produced from the full scan data.
%
%
%
%
%
%
%
%
To account for the spatially varying artifacts, we include the coordinates $(x, y)$, the pattern index $p$ to the inputs, in addition to the measured intensity at each pixel. 
Based on the recent works of \cite{vaswani:2017:attention} and \cite{mildenhall:2020:nerf} in positional encoding, we transform each of $x, y$, and $p$ into 64-length encoded vectors instead of using their original scalar values. 
%
%
%
%
%
Thus, the input to the network consists of a measurement intensity channel, 
and 64 channels each for $x, y$ and $p$, resulting in a total of 193 channels, as shown in Figure \ref{fig:restore_net_arch}. 
The output of the network is a single restored image channel corresponding to the simulated measurement (see Fig.~\ref{fig:restore_net_outputs}(e)).
Details about the network and the dataset are available in section~\ref{sec:network}.



\section{Application: Hyperspectral Imaging}

\begin{figure*}
	\centering
	\includegraphics[width=\textwidth]{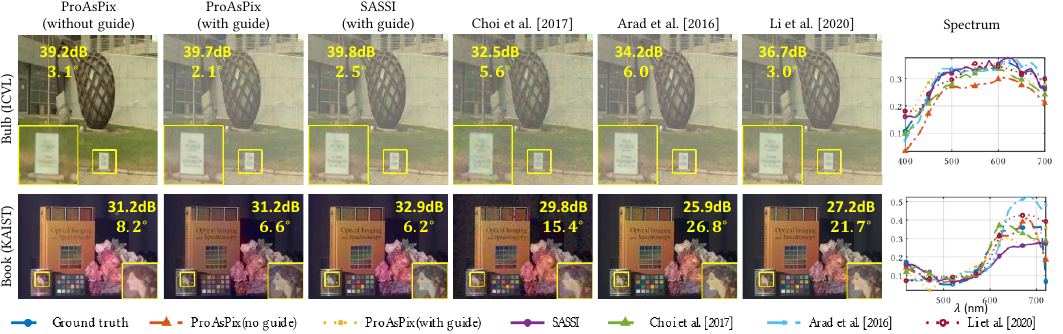}
	\caption{\textbf{Comparisons against snapshot approaches in visible wavelengths.}  We compare ProAsPix, with and without guide, to a number of single-image competitors including spatial coding techniques---SASSI~\cite{saragadam2021sassi} and Choi et al.\ \citep{deepcassi2017}---as well as RGB to HSI techniques \citep{arad_and_ben_shahar_2016_ECCV, Li_2020_CVPR_Workshops}.
Across the board, ProAsPix with and without guide images performs comparable to SASSI and outperforms all other approaches, particularly techniques that recover hyperspectral images from RGB images.}
	\label{fig:snapshot_vis}
\end{figure*}

%
We now consider the application of hyperspectral imaging (HSI) with phase SLMs; our eventual goal is not just to provide a viable framework for HSI but also to quantify the improvement provided by the deep restoration framework.

The image measurements made in \eqref{eq:proaspix01} can be written as, 
\begin{align}
	i_k(x, y) = \int_\lambda h(x, y, \lambda) f^k_\text{SLM}(x, y, \lambda)d\lambda,
	\label{eq:lin_inv}
\end{align}
where $f^k_\text{SLM}$ is an instance of the spatially-varying spectral response of the SLM.
Discretizing and converting images to vectors we get $\bfi_k = X\Phi_k$,
%
%
where $X\in\mathbb{R}^{N_x N_y \times N_\lambda}$ is the matrix representation of the HSI of the scene.
We repurpose the  body of work devoted to HSI reconstruction to solve \eqref{eq:lin_inv} to obtain $h(x, y, \lambda)$ from $\{i_k(x, y)\}$.
%
%

\paragraph{Guide-free reconstruction.}
The traditional approach is to formulate the recovery as a linear inverse problem,
\begin{align}
	\min_{X} \sum_{k=1}^{N}\| \bfi_k - X\Phi_k\|^2 + \mathcal{R}(X),\label{eq:cvx}
\end{align}
where $\mathcal{R}(\cdot)$ is a spatial/spectral regularizer. For snapshot approaches ($N=1$), we used an untrained deep network as a regularizer, similar to the deep image prior (DIP)~\cite{ulyanov2018deep}. We expressed the HSI as the output of a convolutional neural network (CNN) equipped with 2D spatial convolutions and whose input was a fixed noise pattern.
We then solve for the HSI by optimizing for the CNN's parameters. 
%
%
For multi-pattern reconstruction ($N \ge 2$), we found that a combination of 2D total variation (TV) prior and a 1D spectral smoothness prior enabled high quality reconstruction.

\paragraph{Guided reconstruction.}
In the presence of an RGB guide image, we leverage a super pixelation-based reconstruction technique inspired by recent works in hyperspectral imaging~\cite{saragadam2021sassi}.
%
%
Given an RGB image $I_\text{RGB}[x, y]$ of the scene, we first partition the image into $Q$ superpixels~\cite{achanta2012slic}.
We make the assumption that the spectral profiles associated with pixels within any superpixel are scaled multiples of each other.
%
%
Then we model the HSI within each super pixel as a rank-1 matrix, where the spatial component is a scaled version of the grayscale image. 
Specifically, $X_q = \bfg_q \bfs_q^\top$ where $\bfg_q$  is the grayscale image intensity in the $q^\text{th}$ super pixel and $\bfs_q$ is the spectral basis in the $q^{th}$ super pixel.
We then solve for a local least squares problem to estimate the spectral component of the rank-1 matrix, 
%
%
%
%
%
%
\begin{align}
\min_{\bfs_q} \sum_k \|\bfi_{k, q} - \bfg_q \bfs_q^\top\Phi_{k, q}\|^2 + \eta\|\bfs_q\|^2.
\label{eq:guided}
\end{align}
%
%
%
%
%
%
Details about reconstruction in section~\ref{sec:asim}.

\subsection{Simulation}

Figures \ref{fig:snapshot_vis} and \ref{fig:sim_multiframe} provide reconstruction results on simulated data with single and multiple spectrally coded measurements.
Specifically, we simulated the acquisition setup on several datasets including the ICVL~\cite{arad_and_ben_shahar_2016_ECCV} dataset which consists of HSIs with 519 bands over 400 - 1100nm, and KAIST~\cite{deepcassi2017} dataset which consists of HSIs with 31 bands over 420 - 720nm.
%
%
%
We compare our technique against existing snapshot techniques including \citet{deepcassi2017} which consists of a CASSI-type hardware and a deep neural network based reconstruction, and SASSI~\cite{saragadam2021sassi} which consists of a sparse spatio-spectral sampler along with RGB fusion.
We also compared to two techniques that recovered HSIs from RGB images~\cite{arad_and_ben_shahar_2016_ECCV,Li_2020_CVPR_Workshops}.
%

%
%
%
We assume a maximum light level of 1000 photoelectrons a read noise of 2 electrons, resulting in a signal to noise ratio of 30 dB.
%
%
The TV penalty term $\eta_{TV}$ is set according to estimated maximum light level $\tau_\text{max}$ as $\eta_{TV} = \frac{10^2}{\sqrt{\tau_\text{max}}}$, and the spectral penalty term $\eta_\text{spectral}$ is set to $5\times10^{-1}$.
For reconstruction with guided image, the regularization constant $\eta$ in  \eqref{eq:guided} is chosen based on number of measurements with it varying linearly from $10^{-5}$ for a single image to $10^{-6}$ for 256 images.
We quantify performance using peak signal to noise ratio (PSNR) metric and spectral angular mapping (SAM).
%
%
We finally perform guided filtering of the resultant HSI with guide image~\cite{he2010guided} which resulted in up to 2 dB improvement in performance.

\paragraph{Snapshot reconstruction in the visible domain.}
Most HSI reconstruction techniques based on learned models are fine-tuned for the visible domain.
Figure \ref{fig:snapshot_vis} shows reconstruction for HSIs in the visible wavelengths (400 - 700nm) which included SASSI~\cite{saragadam2021sassi} (sparse spatio-spectral + guide image), \cite{deepcassi2017} (SD-CASSI + learned reconstruction), \cite{arad_and_ben_shahar_2016_ECCV} (dictionary-based RGB to HSI), and \cite{Li_2020_CVPR_Workshops} (learned RGB to HSI).
%
%
%
Across the board, ProAsPix results were qualitatively and quantitatively superior.
RGB to HSI techniques \cite{arad_and_ben_shahar_2016_ECCV, Li_2020_CVPR_Workshops} were trained on the ICVL dataset; hence, the RGB to HSI results for the HSI from this dataset (top row)  were comparable to ProAsPix.

\begin{figure}
\centering
	\includegraphics[width=0.475\columnwidth]{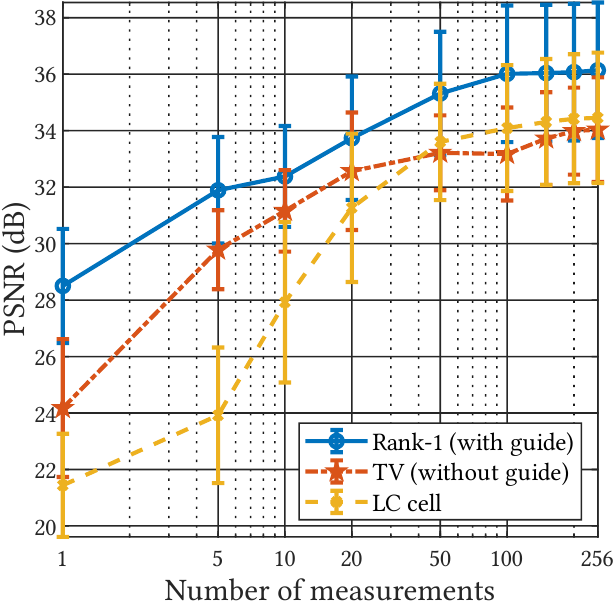}
	\includegraphics[width=0.475\columnwidth]{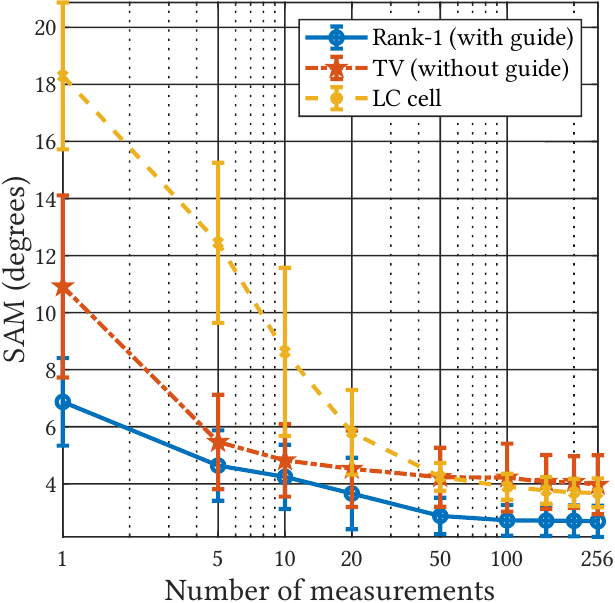}
	\caption{\textbf{Multi-frame performance.} Our proposed approach, with and without a guide image, outperforms the spatially-constant modulation of LC cells, especially when imaging with a small number of image measurements.
	}
\label{fig:sim_multiframe}
\end{figure}

\paragraph{Comparisons with multiple captures.} Our primary competitor for multi frame reconstruction is a single LC cell capture~\cite{oiknine2019compressive} where images are captured with a spatially \textit{invariant} spectral modulation.
Figure \ref{fig:sim_multiframe} plots reconstruction accuracy with LC cell and our technique for varying number of images.
ProAsPix with guided filtering uniformly outperforms the spatially-invariant LC cell.
The reconstruction in the absence of a guide image is better than LC cell at fewer measurements and similar with 50 or more images.
For a small number of images, it is more advantageous to spatially multiplex the various spectral filters.
With a large number of images, spatial multiplexing has a similar effect to capturing images with spatially invariant spectral filters.
%
%
Additional results in section~\ref{sec:ahsi}.

\subsection{Real Results}

\paragraph{Setup.}  We recover HSIs at a spatial resolution of $1024 \times 1024$ and a spectral resolution of $53$ bands in the span of 420 to 940 nm.
Since the spectral filters are linear in $1/\lambda$, we sample these 53 bands uniformly in the reciprocal of the wavelength, which provided a small performance increase  over linear sampling in  wavelength.

\paragraph{HSI reconstruction from a single image.}

\begin{figure}
\centering
\includegraphics[width=0.475\textwidth]{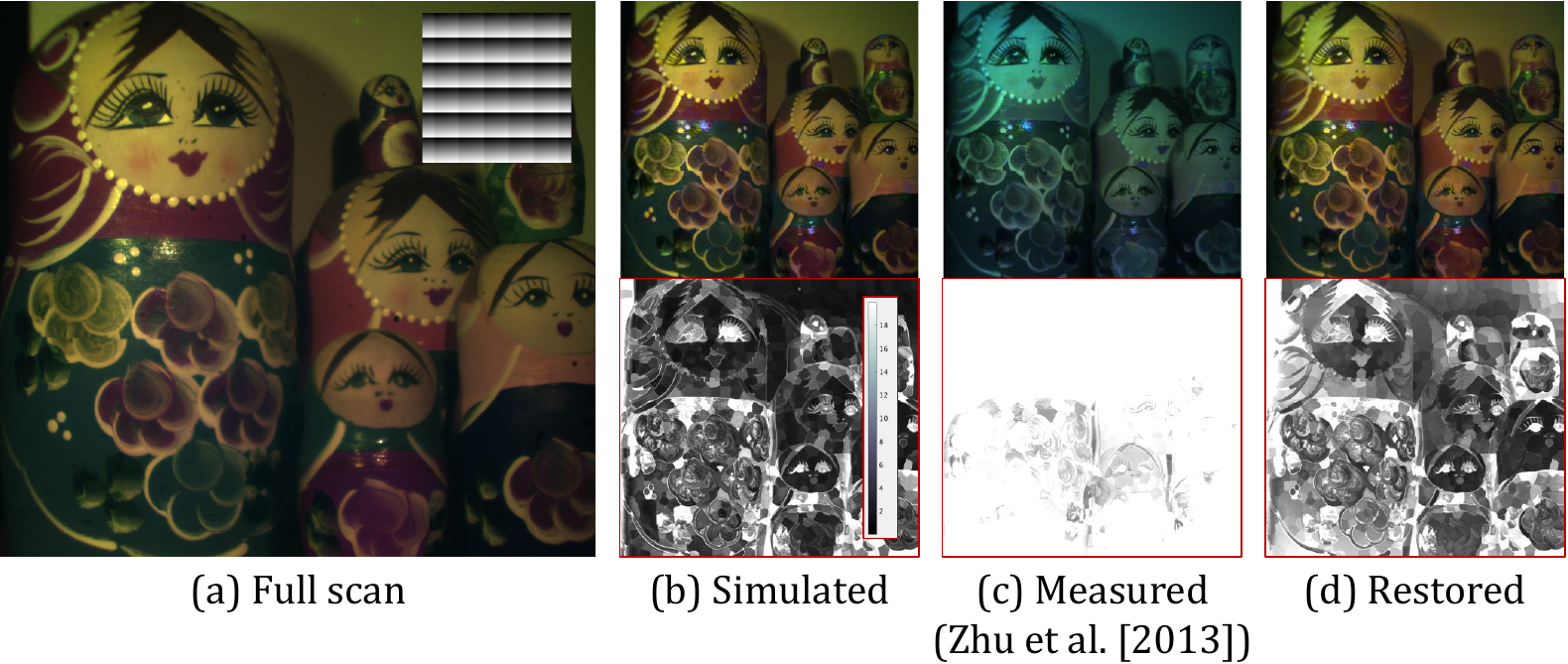}
\caption{\textbf{Comparison of simulated, measured, and restored measurements.} (a) Shown is an RGB image of the scene with inset of the spectral filter array that we used for the results in the other columns. (b-d) We visualize reconstructed HSIs using the simulated, measured (\cite{Zhu:13}), and restored measurements as  rendered RGB images. Below each RGB reconstruction, we provide the angular error against the full scan reconstructions; for these error maps, the brightest values are errors that are $20^\circ$ or higher. }
\label{fig:patternresults3}
\end{figure}

Figure \ref{fig:patternresults3} provides an example of the reconstructions obtained with the simulated, measured, and restored measurements.
Evidently, accounting for the aberrations dramatically improves the performances and produces results similar to ideal simulated conditions.
Next, we observe that 2D patterns generally lead to better reconstructions in real experiments too, for the simulated as well as the restored measurements.
This indicates that the local diversity of patterns does play an important role in reconstruction performance.
%
%
%

\begin{figure}
\centering
\includegraphics[width=0.475\textwidth]{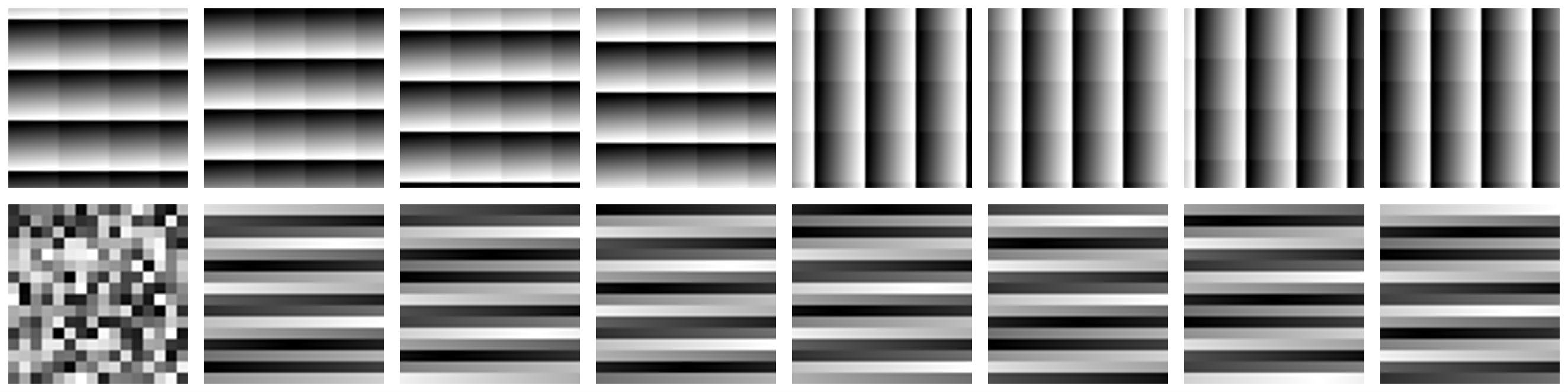}
\caption{\textbf{Patterns used for multi-image hyperspectral imaging.} We greedily chose up to 16 patterns that ensured diversity of spectral filters at each pixel.}
\label{fig:whatpatterns}
\end{figure}

\begin{figure*}
\centering
\includegraphics[width=0.95\textwidth]{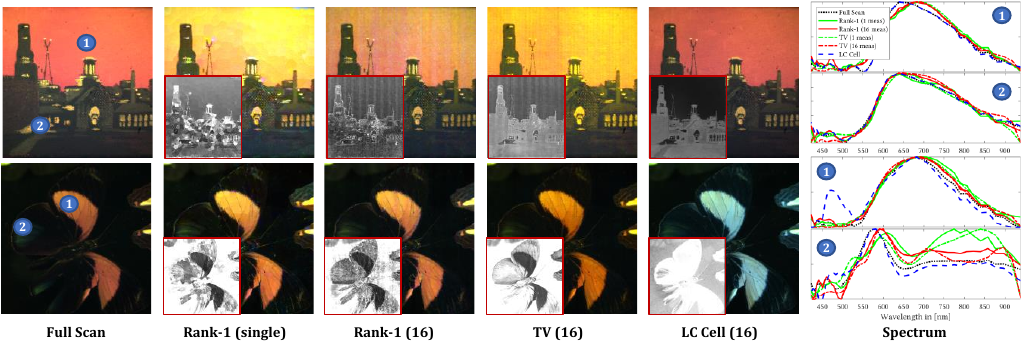}
\caption{\textbf{Qualitative evaluation for multiple patterns reconstructions.} We compare reconstructions from full scan of 256 measurements, a single image reconstruction using the guided Rank-1 technique, and 16 image reconstructions from both the guided rank-1 and guide-free TV techniques. We compare these to reconstructions from an LC cell, with 16 measurements as well. For each scene, we show rendered RGB images and spectrum at two points marked in the first column. Inset on the RGB images are the angular errors, visualized as in other figures with a range of $0$ to $20^\circ$.}
\label{fig:multi}
\end{figure*}

\paragraph{Multi-image reconstructions.} For multi-image reconstructions, we greedily selected a sequence of patterns that provide maximal diversification of filters. 
Starting with the best performing pattern shown in Figure \ref{fig:patternresults3}, which is a 2D horizontal/periodic pattern, we sequentially add patterns that maximizes the addition of new spectral pixel over all previously selected patterns.
The  sixteen such patterns that we select with this scheme is visualized in Figure \ref{fig:whatpatterns}.
We compare the reconstructions with rank-1 guided filter and guide-free TV prior in Figure \ref{fig:multi}.
For the rank-1 approach, we linearly increase the number of superpixels used with the number of measurements.
We also compare to reconstructions that would be obtained with just an LC cell, as opposed to an SLM, which would only provide global spectral modulation.
Overall, ProAsPix works significantly better than what we get with an LC cell.
%
%
%


\section{Other Applications}
\subsection{Spatial Tiling for Material Classification}

\begin{figure}
\centering
\includegraphics[width=0.475\textwidth]{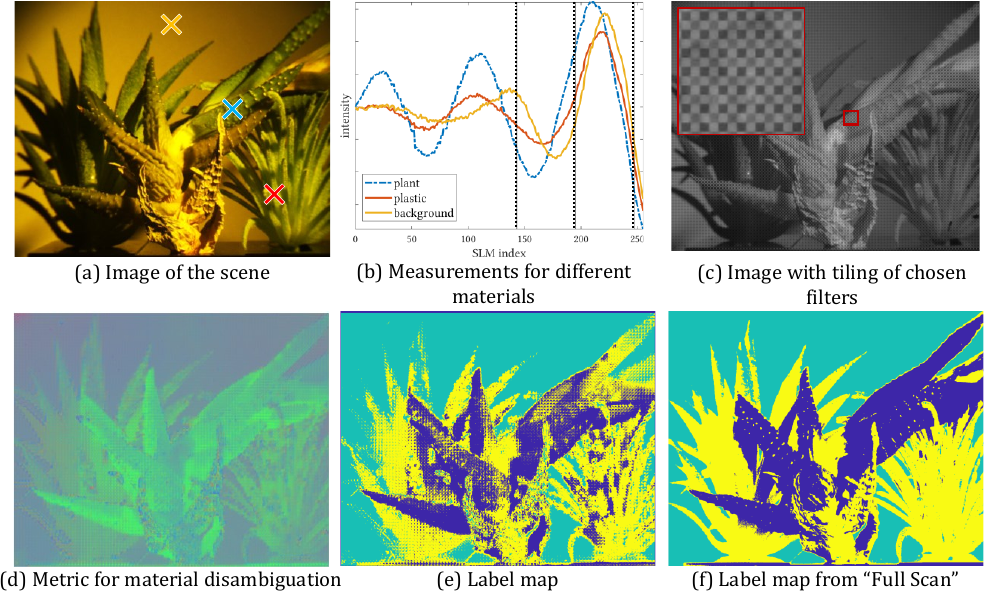}
\caption{\textbf{Disambiguating between materials using programmable spectral filter arrays.} (a) We image a scene with plants, real and plastic. (b) The measurement trace, as a function of SLM input index, is visualized for the two materials as well as the background. We find three index values, marked with dotted vertical lines, that lead to maximally different measurements for the three materials.  (c) An image of the scene is captured with an SLM displaying a checkerboard pattern comprising of the two chosen index values. The inset is the zoomed in version of the cropped region marked in red. (d) Using this single measurement, we can now create a metric that maximally disambiguates between the two materials and threshold it in (e) to get a material map. For comparison, the label map from the full scan  is shown in (f).}
\label{fig:plant}
\end{figure}

\begin{figure*}[!tt]
\centering
\includegraphics[width=\textwidth]{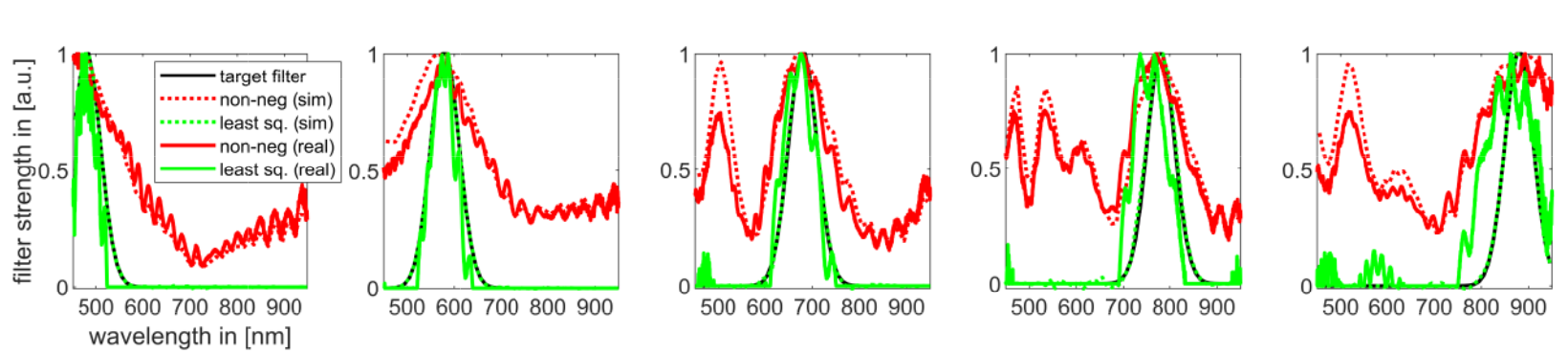}
\caption{\textbf{Implementing arbitrary spectral filters}. The figure shows band pass Gaussian filters with different center frequency, and a standard deviation of $30$ nm. For each target filter (black), we estimate SLM spatial pattern using unconstrained (green) and non-negative least squares fits (red). The plots show both simulated results that provide ideal implementation conditions for this SLM, and real captures using a lab spectrometer. Non-negative least squares provides a poor fit, both in simulation and hardware, likely due to the sinusoidal filter profiles constraining the space of implementable non-negative filters. Unconstrained least square does a significantly better job but is susceptible to noise, especially at deep blur and far red wavelengths, where our light source does not have sufficient spectral radiance.}
\label{fig:progfilter}
\end{figure*}
A powerful capability enabled by our system is that of disambiguating between different materials in a scene.
%
%
%
To obtain an optimal filter array, given a selection of $K$ materials, we first measure their response to the 256 spectral filters that can be created by the SLM.
An example of these measurement traces can be seen in Figure \ref{fig:plant}(b).
We can now select a few, typically a small number of filters, say $Q$ of them, such that measurements associated each of the $K$ materials are distinct.
%
%
We perform a brute-force / greedy scan to identify a set of $Q$ filters that produce the maximal-minimum distance of features on the simplex.
Now that we have our $Q$ filters, we tile them spatially in as compact a block as possible. 
For example, in Figure \ref{fig:plant}(c), we tile $Q = 3$ filters in $2 \times 2$ blocks by repeating one of them to differentiate between real and fake plant.
Once we capture a measurement image, we first demosaick the measurement using standard linear filtering to get a $Q$-channel image.
We project each pixel onto the simplex, and perform nearest neighbor classification on the resulting feature to obtain the material map of the scene (Fig.~\ref{fig:plant}(e)).
%
%
%
%
%
%
This result indicates the immense potential of our system for applying adaptive sensing techniques on top of the setup.

%
%
%


\subsection{Application: Arbitrary Spectral Filters}

An important capability in spectral filtering is the ability to have a programmable filter capable of displaying an arbitrary shape.
LC cells can only implement sinusoidal spectral filters; other results have shown that stacking multiple LC cells offers the ability to get narrowband filters~\cite{lyot1933optical,ohman1938new}.
We explore implementation of spectral filters with profiles that go beyond sinusoids by placing the SLM in the pupil plane, allowing the SLM to modulate light from all scene points.
%
%
%
A key advantage of using SLM is that the switching time between any two filters is same as display rate (60Hz). In contrast, LC tunable filters require hundreds of milliseconds to switch between two wavelengths, which gets larger with increasing distance between the two wavelengths.

Suppose that we seek to implement the spectral filter ${\bf s} \in \reals^{N_\lambda}$.
Let $\Lambda \in \reals^{{N_\lambda} \times 256}$ be the matrix of 256 spectral responses corresponding to different SLM voltages.
Our goal is to solve for weights ${\bf w} \in \reals^{256}$ such that ${\bf s} = \Lambda {\bf w}$; here, the $i$-th element of ${\bf w}$, which we denote of $w_i$ provides the contribution from the corresponding filter.
Once we have ${\bf w}$, we design an SLM pattern which allocates an area corresponding to $w_i$ to the $i$-th index value.

\paragraph{Estimating the weight vector ${\bf w}$.} It is also important the ${\bf w}$ be constrained to be positive since we cannot allocate a negative area in the SLM plane to a filter.
We can estimate such positive patterns by either solving a nonnegative least squares problem, or by solving an unrestricted pattern, and displaying positive and negative parts of the filter separately, capturing two images and subtracting them.
%
%
%
%
%
The latter would require an additional image, but provides a greater space of filters that we can potentially implement. 

\paragraph{Implementing the weight vector on the SLM.}  Given a positive weight vector ${\bf w}$, implementing it on the SLM requires us to divide the SLM area into multiple regions such that the $i$-th region displays the SLM index $i$ and has an area proportional to $w_i$.
We instead  on a simple heuristic for performing this allocation. 
%
%
Next,  SLM pixels in the same column are forces to have the same value, and so we allocate each of the 1024 columns at our disposal to the 256 values proportional to the values in ${\bf w}$.
%
%
Figure \ref{fig:progfilter} shows simulation and real results of implementing bandpass filters.
We generated a set of Gaussian filters with varying center wavelength, and a variance of $30^2\ (nm)^2$.
For each spectral profile, we implemented both standard and non-negative least squares, which are both plotted on top of the ground truth profile.
We can observe that non-negative least square profiles are less precise than that of least squares, for aforementioned reasons.
We also implemented the non-negative filter weights in the lab prototype with a spectrometer in the imaging arm; the results from which are shown in the figure as well.
We observe that non-negative least squares follows the simulations results for the most part.
Least squares offers significantly better fit, both in simulation and in real capture but suffers from photon noise.
%
%
The real spectral profiles oscillate more significantly at the deep blue and far red ends, likely due to poor efficiency of SLM, and low intensity of the incandescent lamp at those wavelengths.

\section{Conclusion}
This paper advances programmable and spatially-varying spectral modulation and discusses its use in multiple applications.
We achieve this capability using an LC-based phase SLM, and develop an optical schematic for implementing it while computationally handling unmodeled aberrations in the setup.
The results in this paper suggest that many of the practical challenges in using phase SLMs for spectral imaging can be computationally addressed in post-processing; this raises intriguing new possibilities in computational imaging and, we hope, opens up the possibility for wider deployment of phase SLMs in spectral imaging problems. 

%

\section*{Acknowledgments}
This work was support in part by a Sony Faculty Innovation Award and the National Science Foundation under the following awards:  1652569, 1730147, and 1801382.

\begin{appendices}
	\section{Basics of Spectral Filtering with LC Cells}\label{sec:lc}

We briefly go over the principle of operation of an LC cell when used to implement a spectral filter.
The reader is referred to \cite{Wu:84} for a detailed treatment of this material.

The basic imaging setup, seen in  Figure \ref{fig:schematics}, consists of an LC cell that is sandwiched between two linear cross polarizers, with their polarization axes oriented at $\pm 45^\circ$ to the LC cell's fast axis.
To describe the propagation of light through this stack, we use  the Jones vector formulation, with the two components   aligned with the fast and slow axes of the LC cell.
Unpolarized light at a wavelength $\lambda$  incident on this filter is first linearly polarized by the first polarizer; since the axis of polarization is at $45^\circ$ to the fast/slow axes of the LC cell --- which we use to define the coordinate axes of a Jones vector, this light can be represented using the Jones vector
\begin{equation}
\frac{1}{\sqrt{2}}  \left[ \begin{array}{c} 1 \\ 1 \end{array} \right].
\label{eq:jones01}
\end{equation}
The birefringence of the LC cell introduces an optical path delay between its fast and slow axes that is equal to
\begin{equation}
\Delta n(v) d_{LC},
\label{eq:lc01}
\end{equation}
where $d_{LC}$ is the thickness of the LC cell, $v$ is the RMS voltage applied across it\footnote{Typically, an LC cell is operated by applying a high-frequency signal to  avoid ion formation and  damage to the device. The specifics of this often controlled by the driver associated with the device. So, when we refer to the control voltage, we refer to the RMS value as opposed to the specific waveform used.}, and $\Delta n(v)$ is the resulting birefringence at this voltage.
This results in a phase difference of 
\begin{equation}
\phi(\lambda) = 2\pi  \frac{ \Delta n(v) d_{LC} }{\lambda},
\label{eq:lc02}
\end{equation}
and hence,  Jones vector after the LC cell is given as
\begin{equation}
\frac{1}{\sqrt{2}}  \left[ \begin{array}{c} 1 \\ e^{j \phi(\lambda)} \end{array} \right].
\label{eq:lc03}
\end{equation}
The second polarizer changes the Jones vector to
\begin{equation}
\frac{1}{2\sqrt{2}} (1-e^{j\phi(\lambda)}) \left[ \begin{array}{c} 1 \\ -1 \end{array} \right],
\label{eq:lc03}
\end{equation}
The intensity of the  light  that exits the filter is given as the square of the magnitude of the Jones vector, which evaluates to
\begin{align}
\frac{2}{8} \left| 1-e^{j\phi(\lambda)}  \right|^2 &= \frac{1}{4} \left( 2 - 2 \cos(\phi(\lambda)) \right) \nonumber \\
&= \frac{1}{2} \left( 1 - \cos\left( 2\pi \frac{\Delta n(v) d_{LC}}{\lambda} \right)\right) \label{eq:lc04}
\end{align}

\begin{figure}
\centering
\includegraphics[width=0.4\textwidth,trim=0 4.1in 0 0, clip]{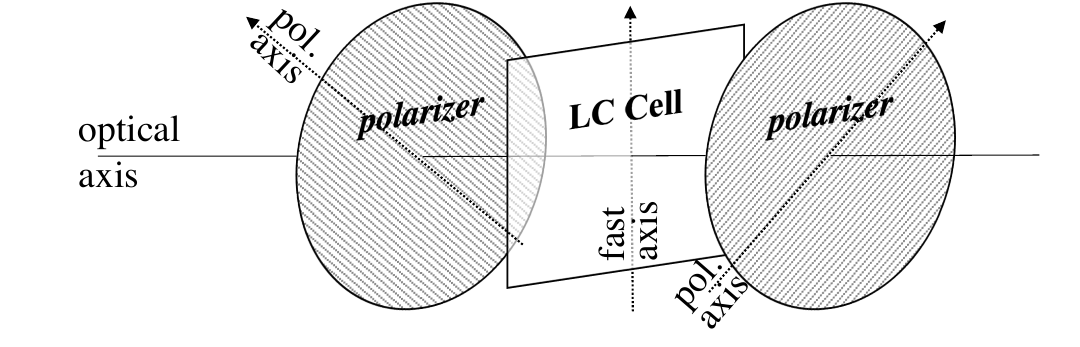}
\caption[]{\textbf{Spectral filtering with LC cells.} An LC cell is placed between two cross polarizers, whose polarization axes are aligned at $\pm 45^\circ$ to the fast axis of the LC cell. This produces a spectral filter whose response can be controlled by changing the birefringence of the cell.}
\label{fig:schematics}
\end{figure}
The expression in (\ref{eq:lc04}) provides spectral filter observed when an RMS voltage $v$ is applied across the LC cell.
This filter is sinusoidal in the wavenumber, or the reciprocal of the wavelength.
The frequency of this sinusoid is determined by the term $\Delta n(v) d_{LC}$,  which quantizes the path difference introduced by the LC cell.
As is to be expected, thicker LC cells  introduce a large path difference which creates spectral filters with more cycles  over $\lambda$.
Similarly, higher levels of birefringence $\Delta n$, that typically happens for low values of $v$, also creates a larger number of cycles over the waveband of interest.

	\section{Details of the Imaging Prototype}\label{sec:hardware}
\paragraph{Specifications of the system.} The system is configured for a waveband spanning  visible  (VIS) and near-infrared (NIR) wavelengths, i.e., 400-1000nm,  which roughly matches  the sensor response of the  main camera that was used for making the spatio-spectral measurements.
We used a Holoeye Pluto2 NIR-015 SLM optimized for NIR wavebands, and endowed with a higher phase retardation which in turn provided a set of spectral filters with larger number of oscillations.
The SLM resolution was $1920 \times 1080$, with a pixel pitch of $8 \mu m$.
The pixel pitch of the camera observing the SLM is $6.5 \mu m$, at a resolution of $2048 \times 2048$ pixels.
The relay lenses used, except for the  one  immediately in front of the camera(s), were achromatic doublets, optimized for NIR transmissions, which we observed provided a smaller spot size over the VIS-NIR wavebands.
%
%
However, the use of doublets does lead to a chromatic blur in our captured images, which affects the captured imagery.
Clearly, using better optics, perhaps apochromatic lenses optimized to provide better chromatic correction, would naturally lead to better results.
The lab prototype above had a spatial field of size approximately $10.4 \times 8$ sq.mm; this field size, that is smaller than the dimensions of the SLM or camera, allowed us to avoid strong spherical and chromatic aberrations at its  periphery.
We also set the Fourier plane aperture to be 10mm wide; with 100mm relay lens,  this resulted in a  system with an $f/10$ aperture.

\begin{figure}
	\centering
	\includegraphics[width=0.475\textwidth]{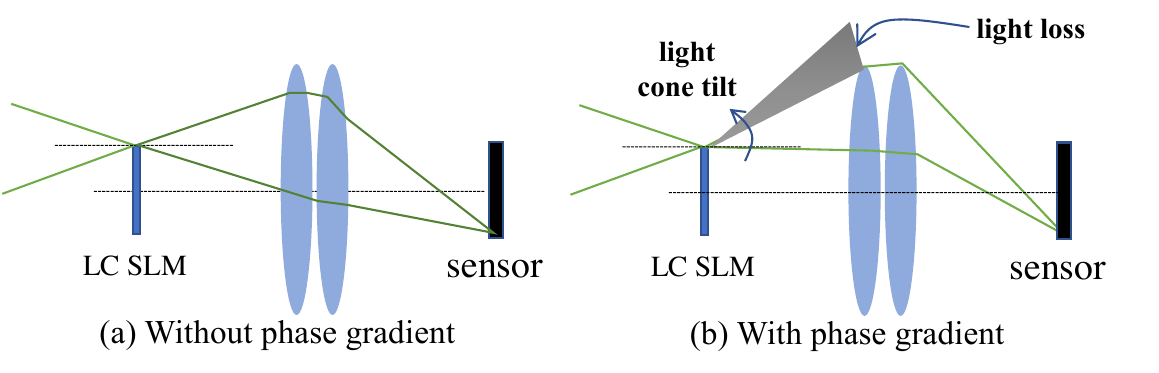}
	\caption{\textbf{Vignetting due to phase modulation.}  As compared to the scenario where a constant phase pattern is shown on the SLM, as seen  in (a), a spatially-varying phase retardation distorts the wavefront by introducing local tilts, as seen in (b). This tilt can lead to a portion of the incident light being blocked by the relay lenses.}
\label{fig:lightloss}
\end{figure}

\paragraph{Unintended phase modulation.} A complicating factor in that we subject the light incident on the SLM to a spatially-varying phase retardance, which distorts the incident wavefront
Such  distortions introduce local tilts to the incident light that is dependent on the spatial gradient of the phase pattern, which  can have undesirable effects due to non-idealities in the optical relay between the SLM and the camera.
%
%
%
%
To see why this is the case, consider the light path between the SLM and the image sensor, shown in Figure \ref{fig:lightloss}.
When there is no phase gradient on the SLM, i.e., a spatially-constant phase as is the case with an LC cell, we get the scenario in Figure \ref{fig:lightloss}(a), where there is no light modulation or loss except for the spectral filtering.
Now consider the scenario in Figure \ref{fig:lightloss}(b) where due to the presence of the phase gradient, we get a local tilt of the wavefront.
A point on the top of the SLM with an unfavorable tilt could result in part of the incident cone being tilted beyond the aperture of the relay lenses.
%
%
As a consequence, the change in intensity that we observe at a sensor pixel, when displaying a spatially-varying pattern on the SLM, ends up being a complex function of spatial location as well as phase gradient. 
We next look at approaches for reducing these undesirable effects of phase modulation.

\paragraph{Non-ideal optics.} When imaging with broadband illumination, it is common to observe field aberrations in the form of large focus spot sizes and spatially-varying distortions.
However, this means that the mapping between SLM and sensor pixels in no longer one-to-one; since, each sensor pixel gathers light from multiple SLM pixels, there is a blurring of the spectral filter that we would like to implement at the pixel.

\paragraph{Discontinuities in the displayed pattern.} The retardance in an SLM, while spatially-varying, is constrained to be smooth due to physics of the construction of the SLM.
This naturally results in strong distortions near sharp changes in the SLM pattern.
To mitigate these non-idealities in the implementation, we provide two complementary approaches: the first dealing with bounding the gradient of the displayed pattern to reduce phase distortions, and the second, to computationally restore the captured measurements.


\subsection{Calibration}

The proposed system requires  key calibration steps that need to be performed for correct operation.
First, we need to measure the spectral filter created by the LC cells under different input voltage, or what we refer to as spectral calibration.
And second, we need to get a pixel-to-pixel mapping between the SLM and the camera(s); this step is especially important since our object is to implement a filter array and hence, this $XY$ calibration provides critical information on the filter applied to obtain the measurement at a sensor pixel.


\paragraph{Spectral calibration. } We perform spectral calibration by mounting a spectrometer (OceanOptics FLAME-S-VIS-NIR-ES)  in place of the  main camera and sweep $256$ constant-valued patterns on the SLM, corresponding to its 8-bit input. 
For each displayed pattern, we capture a spectral measurement over the range $[380, 1010]nm$ at a spectral resolution of $0.35nm$.
Recall, that the actual voltage across the LC cells in the SLM is determined by the gamma curve.
%

\begin{figure}
\includegraphics[width=0.475\textwidth]{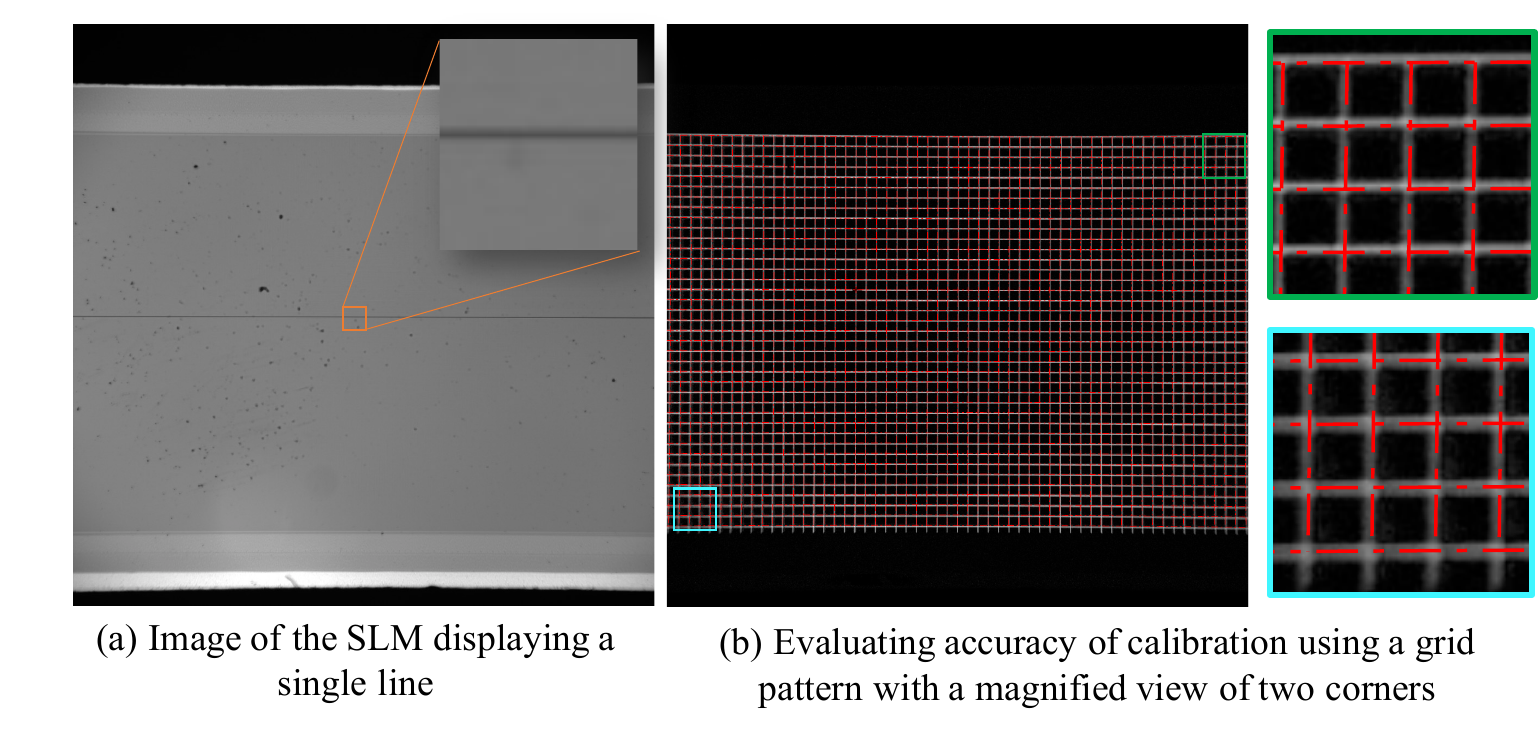}
\caption{{\textbf {XY calibration.}} (a) The SLM is calibrated by first performing a row  and column scan, where an individual row/column displays a different value from the rest. We use this to fit a third-order polynomial between the camera pixel indices to that of the SLM.  (b) We use this grid pattern to refine / adjust the calibration parameters. The use of a grid enables us to correct for drift in the calibration by capturing just a single image.}
\label{fig:xycalib}
\end{figure}

\paragraph{SLM-to-camera alignment.} To recover hyperspectral images from the measurements acquired with the main camera, we need the knowledge of which SLM pixel is observed at each camera.
With ideal optics, this mapping would be given by a homography.
However, due to the presence of radial distortions in the image, we needed to use a more complex model for the fit.

To get the SLM to sensor mapping, we captured a collection of images, with the device imaging a diffuser that was uniformly lit with a broadband light source.
Each captured image was of the SLM displaying an all-zero pattern, except on a single row or column that was set to $255$.
Figure \ref{fig:xycalib}(a) provides an example of such an image, where the SLM displays a single row.
We detect the location of the line on the sensor coordinates using simple comparison against a reference capture with  all-zero pattern on the SLM.
After some pre-processing, we obtain a collection of pixels on the sensor that maps to this particular row on the SLM.
The SLM sweeps through multiple  rows, staggered by a small amount --- 33 pixels in our implementation.
At the end of this row scan, we end up with a collection of $N$ correspondences of the form $\{ x_c^i, y_c^i, y_s^i \}_{i=1}^N$, where the sensor locations $(x_c^i, y_c^i)$ map to SLM row $y_s^i$.
We fit a third-order polynomial that maps $(x_c, y_c)$ to $y_s$, using RANSAC to obtain a robust fit.
The same process is repeated for columns of the SLM and, as with the rows, we obtain a third-order polynomial fit.

Once we have the polynomial mapping from the sensor to the SLM, we invert this mapping by generating a large number of point-to-point correspondences, and applying the polynomial fit for the rows and columns in reverse --- thereby obtaining two cubic polynomials that relate $(x_s, y_s)$, a point on the SLM, to $x_s$ or $y_s$, the corresponding row/column on the sensor.
We visualize the accuracy of this fit in Figure \ref{fig:xycalib}(b), where we display a grid pattern on the SLM --- with a spacing that is different from what was used previously in row/column sweep  --- and overlay the predicted grid location on top of the sensor image.
We generally observe an accurate match between the measured and observed grid, even at the very boundaries of the captured image.
Quantitatively the average error in calibration is approximately 0.1 pixels, with a maximum error of less than 0.5 pixels.

We also use the grid image shown in Figure \ref{fig:xycalib}(b) to incrementally correct a pre-calibrated system using just a single capture, along with an all-zero reference for comparison.
Such a correction, applied periodically, allows us to account for small displacements due to mechanical vibration.

\paragraph{Guide alignment.} Aligning the guide RGB camera to the main camera was done by imaging a highly-textured scene with the SLM displaying a constant pattern, and hence, no spatially-varying filters.
We then used standard image registration techniques by detecting SURF feature points and descriptors~\cite{bay2006surf}, establishing correspondences, and fitting a homography model to it.

\begin{figure}
	\centering
	\includegraphics[width=0.4\textwidth]{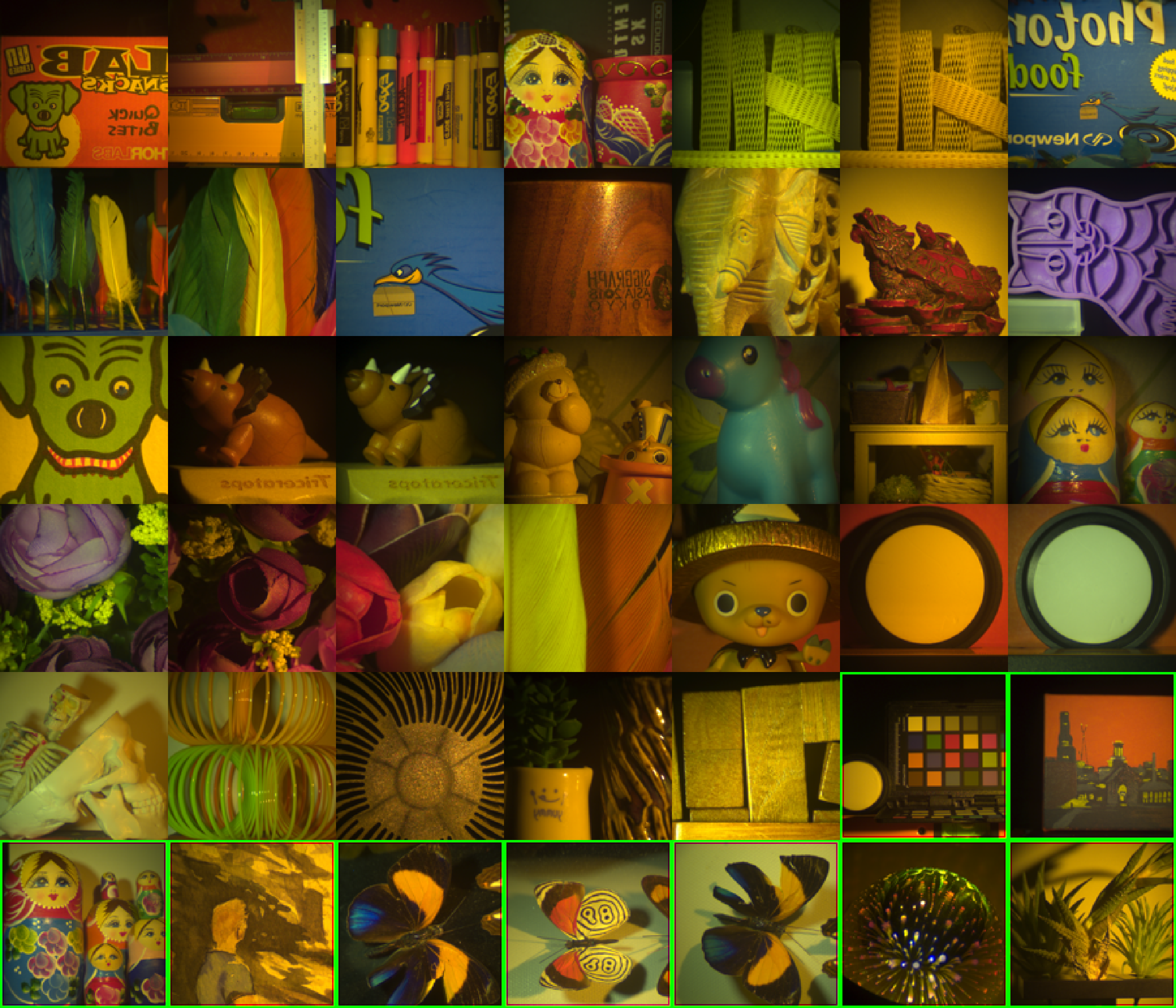}
	\caption{\textbf{Dataset.} We capture of dataset of 42 scenes. For each scene, we capture a Nyquist scan of 256 images, corresponding to constant patterns displayed on the SLM, and the 92 spatially-varying patterns as described in  Figure \ref{fig:pat2}. The scenes marked with a green boundary were used for testing and the rest for training the restoration network.}
	\label{fig:dataset}
\end{figure}

\begin{figure}
	\centering
	\includegraphics[width=0.475\textwidth]{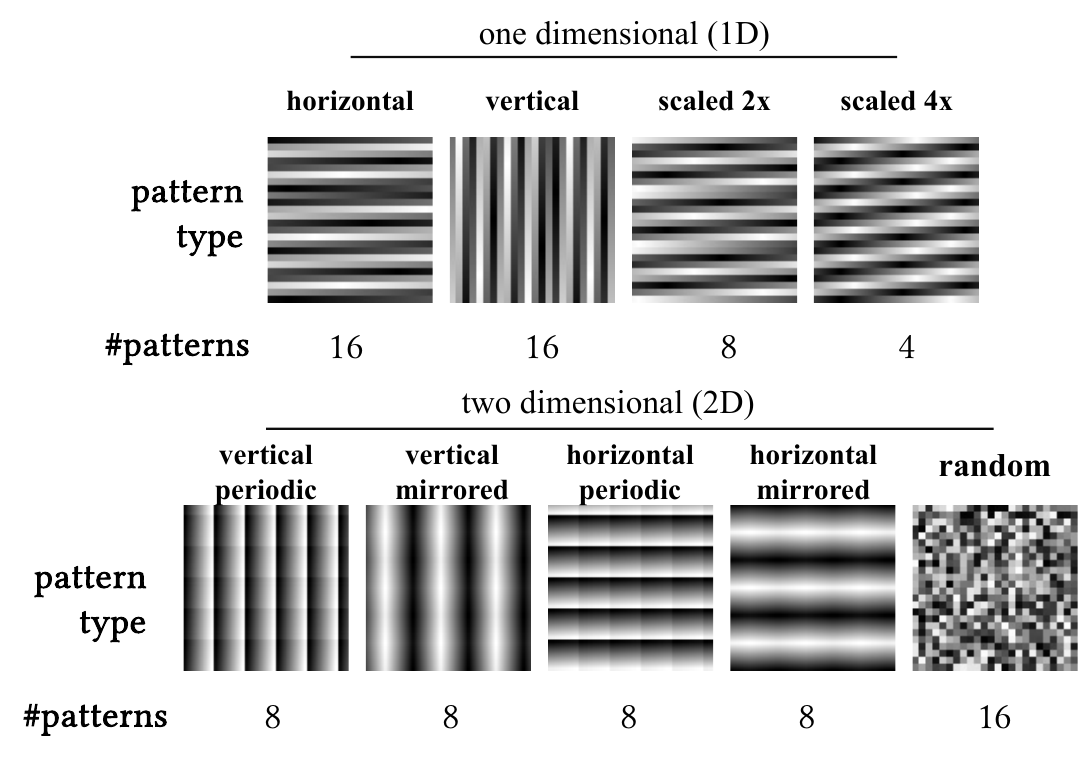}
	\caption{\textbf{SLM patterns used in training.}  For each scene in our dataset, we capture 92 image measurements by showing spatially-varying patterns on the SLM. Shown here are the set of patterns used which comprises of one-dimensional patterns with staggered horizontal and linear patches, and two-dimensional patterns with linear ramps, with periodic and mirrored tilings. We create multiple patterns of each type by shifting them spatially, except for the random patterns where we simply regenerate them.}
	\label{fig:pat2}
\end{figure}

\paragraph{Dataset.} We created a dataset of 42 scenes using our lab prototype. 
The dataset comprised of indoor scenes, comprising mainly of single or multiple objects.
Figure \ref{fig:dataset} provides the RGB image, captured with the guide camera, for each of the scenes.
We illuminated the scenes with a number of different sources including an NIR-enhanced incandescent light, a cool white LED, and CFL lamps.
All acquired images, both from the grayscale camera and the RGB camera, are registered to the SLM using the calibration.
After mapping to the SLM, the images are cropped to the central $1024 \times 1024$ pixels.
For each scene, we captured the set of 256 ``full scan'' patterns and a set of 92 spatially-varying SLM patterns.
The 92 spatially varying patterns are shown in Figure \ref{fig:pat2} and they balance between enabling a smooth tiling, so as to minimize aberrations when implemented on the SLM, to having rich local diversity, to enable single shot reconstruction.

\begin{figure}
	\centering
	\includegraphics[width=0.475\textwidth]{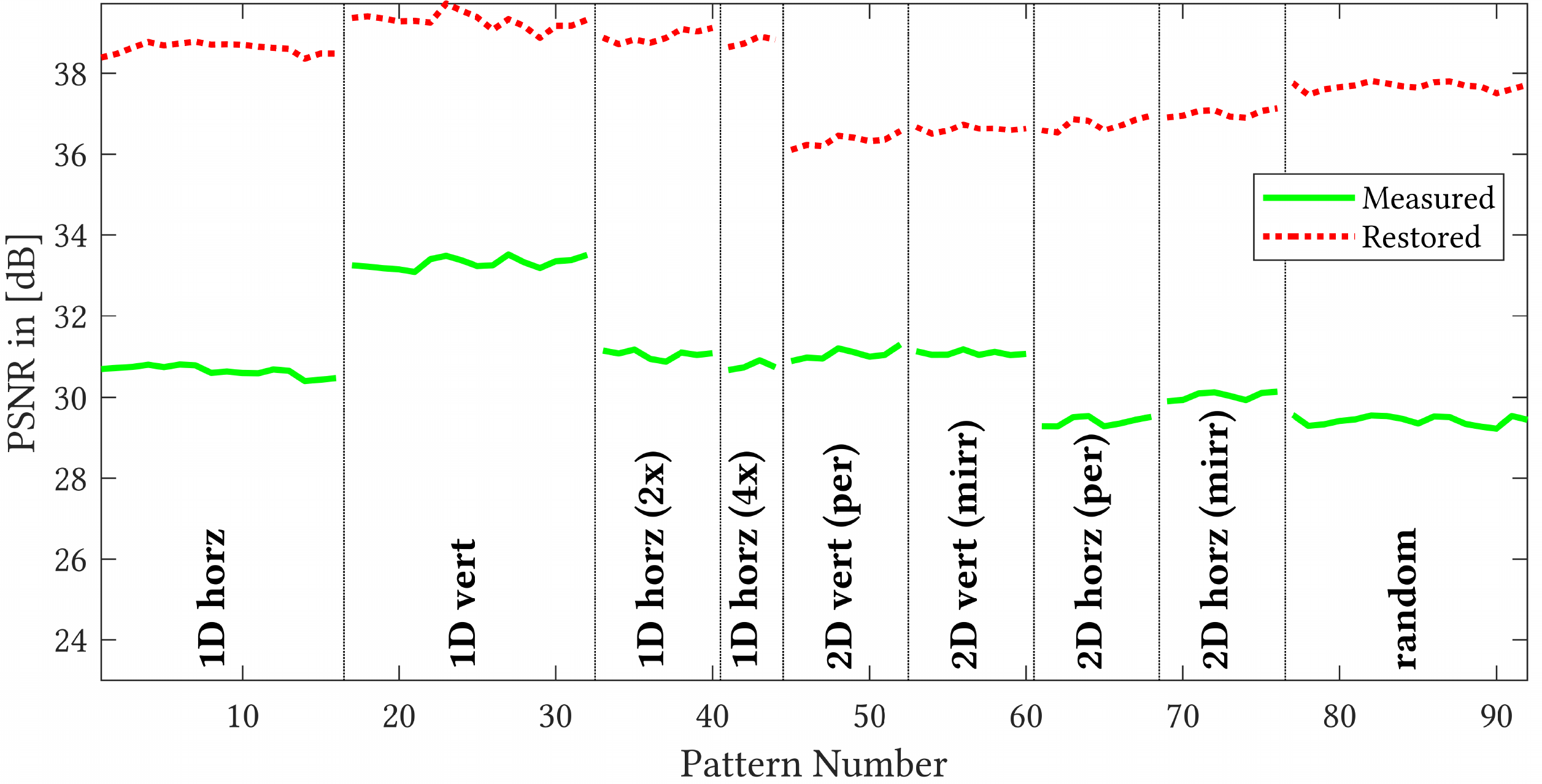}
	\caption{\textbf{Improvements from restoration.} The restoration network provides measurements that have a higher fidelity to the ideal model, by correcting for blur and vignetting introduced by impreciseness in the optical implementation. We can see that, across all 92 SLM patterns, it provides an improvement of 7 dB or higher, as compared to the raw measurements. The numbers shown here are averages, over all the test scenes.}
	\label{fig:restore_patterns}
\end{figure}

\paragraph{Performance of the restoration strategy.} Figure \ref{fig:restore_patterns} evaluates the performance of the restoration network on a test set of seven scenes. 
We quantify the error between the simulated and restored measurements, averaged across the seven scenes, for each of the 92 patterns, and compare them to that of the error associated with the measured images.
We use peak signal-to-noise ratio (PSNR), measured in dB, as our choice of metric which is defined as:
\begin{equation}
	\textrm{PSNR }  = 20 \log_{10} \left( \frac{ \| \bfx \|_\infty}{ \textrm{RMSE}({\bf x}; \widehat{{\bf x}})  } \right),
\end{equation}
where ${{\bf x}}$ and ${\widehat{{\bf x}}}$ are the simulated image and the restored/measurement image, respectively, with an intensity range [0--1].
We observe an improvement that is often greater than 
7 dB, 
indicating significant increase in measurement fidelity to the simulated data once we apply the restoration network.

	\section{Designing the Gamma Curve}\label{sec:gamma}

\begin{figure*}
\centering
\includegraphics[width=\textwidth] {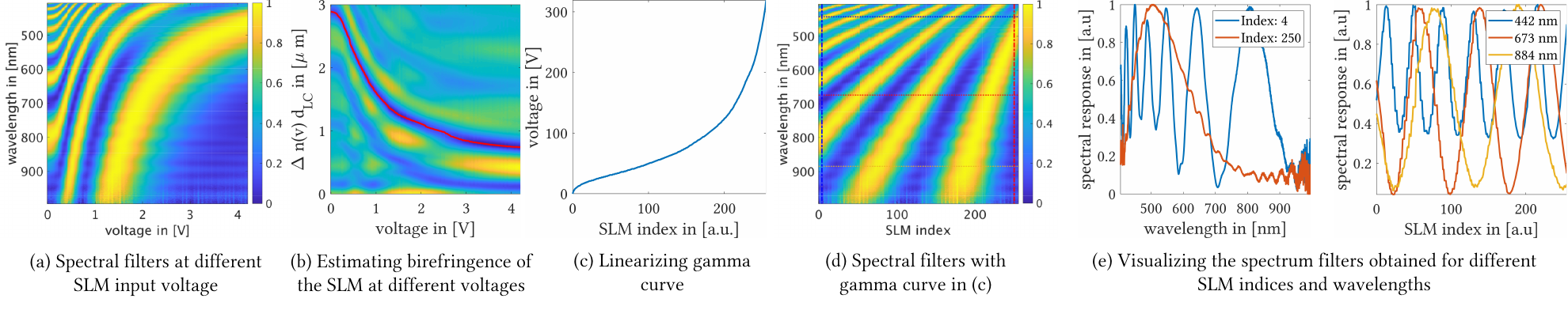} 
\caption[]{\textbf{Procedure for designing the SLM gamma curve.} Our goal is to estimate a gamma function such that the nonlinear relationship between SLM birefringence and input voltage is linearized. To do this, we use a spectrometer to measure the spectral filter produced over a range of input voltages. The measured data is shown in (a). For each voltage $v$, we brute-force search for the value of $\Delta n(v) d_{LC}$ that fits the spectral filter obtained at that voltage by measuring accuracy to (\ref{eq:lc04}). The resulting loss function is visualized as an image in (b). Identifying the minimum for each voltage provides us with value of $\Delta n(v) d_{LC}$, overlaid as a red curve in (b). We use this estimated value to design a gamma curve $\gamma(\cdot)$  for the SLM, shown in (c), such that the overall function $\Delta n(\gamma(\cdot)) d_{LC}$ is linear. (d, e) The resulting set of filters that we obtain, now indexed as a function of the 8-bit index used in controlling the SLM.}
\label{fig:gamma}
\end{figure*}

The construction of the gamma curve requires knowledge of the term $d_{LC}   \frac{\partial \Delta n}{d \gamma}$.
This depends on the term $d_{LC} \Delta n(\cdot)$, which we estimate using the following procedure.

\textit{Step 1 --- Spectral filters as a function of input voltage.}  We place a spectrometer in place of the sensor in Figure \ref{fig:schematics}, and sweep a constant pattern at known voltage on the SLM.
For the SLM that we used, which we describe in more detail later, the working range with best diversity of spectral filters was $[0, 4.2] V$, and hence, we linearly scanned the SLM in this range and measured the resulting spectral filter for each voltage.
The resulting collection of spectral filters is visualized in Figure \ref{fig:gamma}(a).

\textit{Step 2 --- Estimating $\Delta n(v) d_{LC}$ as a function of input voltage $v$.}
In an ideal setting, the spectral filter observed would be a perfect fit to (\ref{eq:lc04}) and hence, we can estimate the value of $\delta n(v) d_{LC}$ for each input voltage $v$ analytically.
However, in practice, non-idealities in the the SLM as well as optics used produces deviations in the spectral filter.
So, we perform a brute force search over a range of values --- in this case, $[300, 3000] nm$ --- and finding the value that has the least misfit to the analytical expression given in (\ref{eq:lc04}).
This procedure provides us with an empirical estimate of $\Delta n(v) d_{LC}$, as shown in Figure \ref{fig:gamma}(b).

\textit{Step 3 --- Estimating $\gamma$.} Our goal is to estimate a function $\gamma$ such that $d_{LC} \Delta n(\gamma(\cdot))$ is linear in its argument.
Recalling that our SLM is controlled via a 8-bit video signal, there are 256 possible inputs to the $\gamma$ function which maps to a voltage value that is applied to the SLM.
We also need to ensure that this gamma function preserves the range of voltages such that $\gamma(0) = 0V$ and $\gamma(255) = 4.2V$.
To achieve this we sample 256 uniformly-placed values between the $\Delta n(0 V) d_{LC}$ and $\Delta n(4.2 V) d_{LC}$, the maximum and minimum values taken.
For each value, we find the voltage $v$ that achieves the specified birefringence.
This value of voltage is what the $\gamma$ function maps onto, for each of its 256 inputs.
This resulting $\gamma$ for our SLM is visualized in Figure \ref{fig:gamma}(c).

 As noted earlier, this choice of $\gamma$ makes the phase gradient solely dependent on the  spatial gradients of the pattern $p(x, y)$ that we display on the SLM.

	\section{Network Design and Training}\label{sec:network}
	The number of channels when passing through the downsampling blocks are 192, 384, 768, and 768, and the upsampling blocks have them in reverse. 
Each downsampling block comprises of a convolution layer of kernel size $3 \times 3$ followed by a convolution layer of kernel size $4 \times 4$ with two strides for the downsampling operation. 
Each upsampling block comprises of a convolution-transpose layer of kernel size $4 \times 4$ with two strides for upsampling operation followed by a convolution layer of kernel size $3 \times 3$. 
The input is first passed on to an input convolutional layer of kernel size $3 \times 3$, and then on to the downsampling and upsampling blocks, and finally to an output convolutional layer of kernel size $3 \times 3$. 
All convolutional layers use a reflection padding of one, and all use Leaky ReLU activation with a negative slope of 0.2 except the output layer which uses none.

\paragraph{Training.} We use the measurement and simulated images of 15 scenes, each with multiple SLM patterns, as our training set. 
We randomly pick $64 \times 64$ patches from measurement and simulation images as training input-output pairs. 
%
Similarly, we use patches from measurement and simulated images of eight other scenes for validation.
We use L2-norm of the difference between the estimated and ground-truth images as our loss function to be minimized. 
We avoid image priors such as gradient sparsity since our operative images are not ``natural.'' 
We use Adam optimizer \cite{kingma:2015:adam} with a learning rate of $10^{-3}$, $\beta_1 = 0.9$, and $\beta_2 = 0.999$. 
We train the network for 100k iterations with a batch size of 500 that takes around 45 hours using four Titan Xp GPUs.

	\section{HSI Reconstruction Algorithms}\label{sec:ahsi}

\subsection{Problem Formulation}
Each spatial pixel of the SLM implements a spectral filter governed by its retardance.
%
%
Rewriting the output of SLM in terms of spectral filter at each location gives us,
\begin{align}
	i_k(x, y) = \int_\lambda h(x, y, \lambda) f^k_\text{SLM}(x, y, \lambda)d\lambda,
\end{align}
where $f^k_\text{SLM}$ is an instance of the spatially varying spectral response of the SLM.
Discretizing the equation gives us,
\begin{align}
	I_k[m, n] = \sum_{l=1}^{N_\lambda}H[m, n, l]\Phi_k[m, n, l].
\end{align}
Converting $I_k[x, y]$ into a vector $\bfi_k \in \mathbb{R}^{N_xN_y \times 1}$, we can express the measurement model as
\begin{align}
	\bfi_k = X\Phi_k,\label{eq:lin_inv}
\end{align}
where $X\in\mathbb{R}^{N_x N_y \times N_\lambda}$ is the matrix representation of the HSI.
Equation \ref{eq:lin_inv} shows that the measurements are a linear transform of the scene's HSI, and hence can be recovered using linear inverse approaches.
We now describe the two algorithms mentioned in the main paper to solve for the scene's HSI based on the availability of an extra RGB buide image.

%
%

\subsubsection{Reconstruction without using Guide Image}
The simplest way to solve for the scene's HSI is to formulate a convex optimization problem:
\begin{align}
	\min_{X} \sum_{k=1}^{N}\| \bfi_k - X\Phi_k\|^2 + \mathcal{R}(X),\label{eq:cvx}
\end{align}
where $\mathcal{R}(\cdot)$ is a spatial/spectral regularizer.
Classical choices for regularizers involve 2D total variation (TV) prior for the spatial dimensions and 1D smoothness prior for spectral dimension. Combining them gives us,
\begin{align}
	\min_{X} \sum_{k=1}^{N}\| \bfi_k - X\Phi_k\|^2 + \eta_{TV} TV_{2D}(X) + \eta_\text{spectral}\|XD_\lambda\|^2,
	\label{eq:tv}
\end{align}
where $D_\lambda$ is the 1D difference operator acting along spectral dimension.

A different form of regularizer would be to leverage inductive biases of randomly initialized neural networks.
This approach, called deep image prior (DIP)~\cite{ulyanov2018deep} has shown promising results in several imaging tasks such as denoising, super resolution and in painting.
We can leverage DIP for solving eq.~\ref{eq:cvx} as follows.
Let $\mathcal{N}_\theta$ be a randomly initialized convolutional neural network (CNN)  equipped with 2D convolutions. The network takes fixed noise, $Z$ as input and outputs a multichannel image -- in this case, the hyperspectral image $X$. 
The goal is then to optimize for the parameters of the network $\theta$ by solving the following optimization problem,
\begin{align}
	\min_{\theta} \sum_{k=1}^{N}\| \bfi_k - \mathcal{N}_\theta(Z)\Phi_k\|^2.
\end{align}
The DIP formulation leverages the inductive biases of CNNs to implicitly regularize the linear inverse problem, thereby leading to high quality reconstruction.
Empirically, we found that the DIP formulation led to higher accuracy when the number of captured images is $N=1$.
For multi-pattern recovery ($N > 1$), TV-based formulation performed as well as DIP-based formulation.

\paragraph{Implementation details.} We solved both TV-based and DIP-based recovery approaches using stochastic gradient descent approach in PyTorch~\cite{pytorch2019}. We used Adam~\cite{kingma2014adam} optimizer for this purpose with a learning rate of $10^{-2}$ and $\beta_1=0.9, \beta_2=0.999$ for running averages of gradients and square of gradients, respectively. The optimization was run for a total of $200$ iterations.
The values of $\eta_{TV}$ and $\eta_\text{spectral}$ were estimated empirically based on the number of captured images and provide a trade-off between robustness to noise and data fidelity.

\begin{figure}[!tt]
	\centering
	\includegraphics[width=\columnwidth]{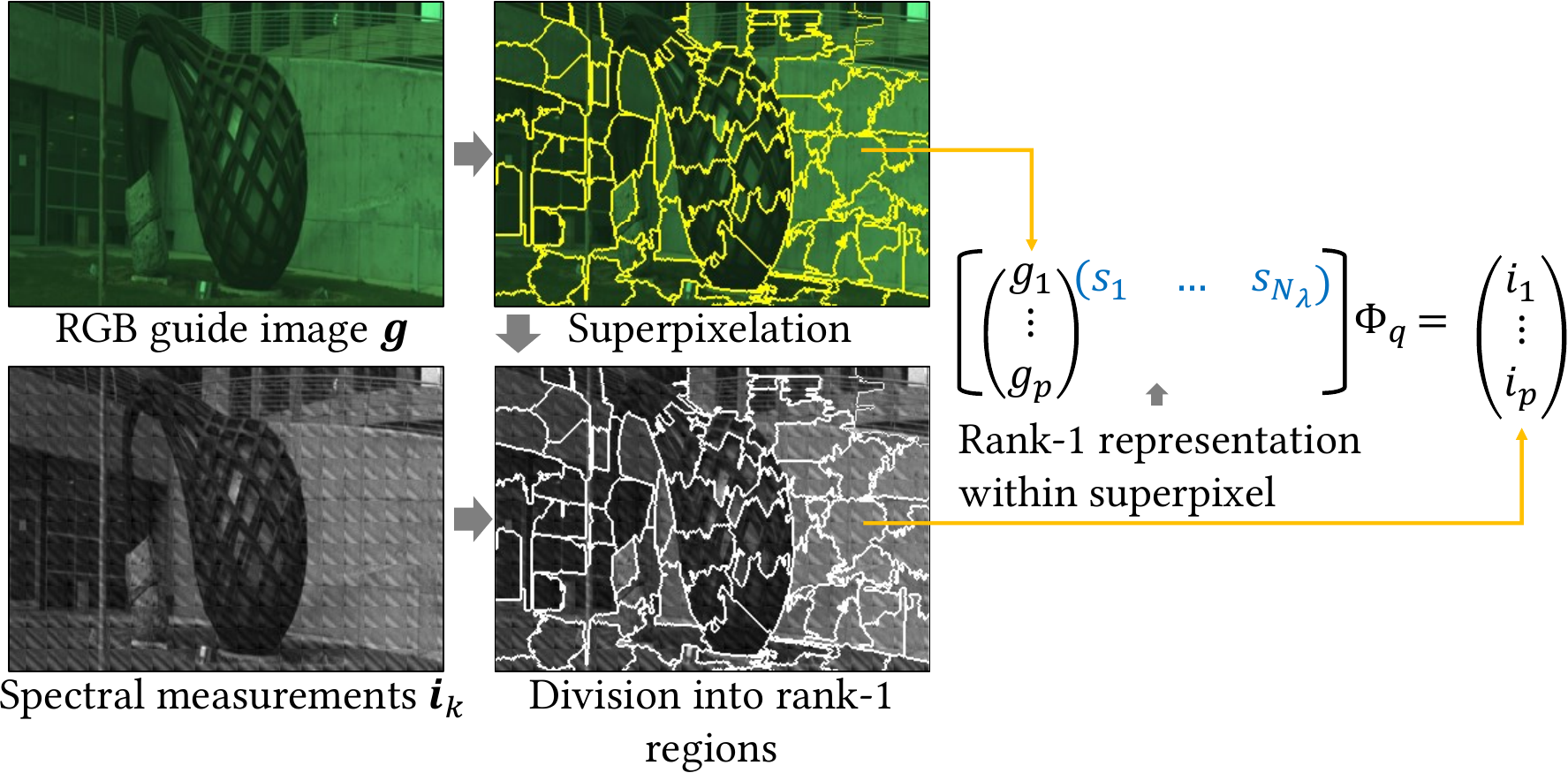}
	\caption{\textbf{Rank-1 reconstruction with guide image.} In the presence of a guide image, we exploit spatial homogeneity to get accurate reconstruction. We divide the scene into homogeneous regions via superpixelation. We then represent the HSI within each superpixel as a rank-1 matrix, allowing us to robustly obtain the spectrum for each superpixel.}
	\label{fig:rank1}
\end{figure}

\subsubsection{Reconstruction using Guide Image}
The availability of an extra guide in the form of an RGB image can significantly increase the reconstruction accuracy.
While it is possible to consider the RGB image as another linear measurement of the scene's HSI and solve a joint optimization problem, we follow a non-linear approach that provides a better estimate.
Our guided reconstruction technique is inspired by recent work in hyperspectral imaging~\cite{saragadam2021sassi}, where superpixels are used to create a scene specific prior.

Given an RGB image $I_\text{RGB}[x, y]$ of the scene, we first partition the image into $Q$ superpixels using the simple linear iterative clustering (SLIC)~\cite{achanta2012slic} algorithm.
We make the assumption that the spectral profiles associated with pixels within any superpixel are scaled multiples of each other.
The underlying motivation for this approach is that superpixels adhere to object and depth boundaries and hence partition the image into regions of homogenous color, and thereby homogenous spectra.

We explain the recovery technique for one super pixel -- the extension to all super pixels is straightforward.
Consider the $q^{th}$ super pixel. 
Let $\bfg_q$ be the grayscale pixel intensity of the guide image, $\bfi_{k,q}$ be the spectrally filtered measurements in the $k^{th}$ image, $\Phi_{k, q}$ be the measurement matrix corresponding to the $k^{th}$ image in the $q^{th}$ superpixel.
Since we assumed that the spectral information is identical up to scale in this superpixel, we rely on a simple rank-1 model to represent the HSI of the superpixel.
Specifically, 
\begin{align}
	X_q &= \bfg_q \bfs_q^\top,
\end{align}
where $\bfs_q$ is the spectral basis in the $q^{th}$ super pixel.
We can now solve the following regularized linear inverse problem at each superpixel using a simple least squares solution:
\begin{align}
\min_{\bfs_q} \sum_k \|\bfi_{k, q} - \bfg_q \bfs_q^\top\Phi_{k, q}\|^2 + \eta\|\bfs_q\|^2.
\label{eq:guided}
\end{align}
%
An overview of this approach is visualized in Fig.~\ref{fig:rank1}.
The number of superpixels is dependent on the noise levels and number of images.
As a general rule, the higher the noise, the fewer the superpixels so as to have a stable inversion.
Similarly, the more the images, the smaller the superpixels.
%


	\section{Additional Real and Simulation Results}\label{sec:asim}
	
\subsection{Simulations}

\begin{figure}[!tt]
	\centering
	\includegraphics[width=0.8\columnwidth]{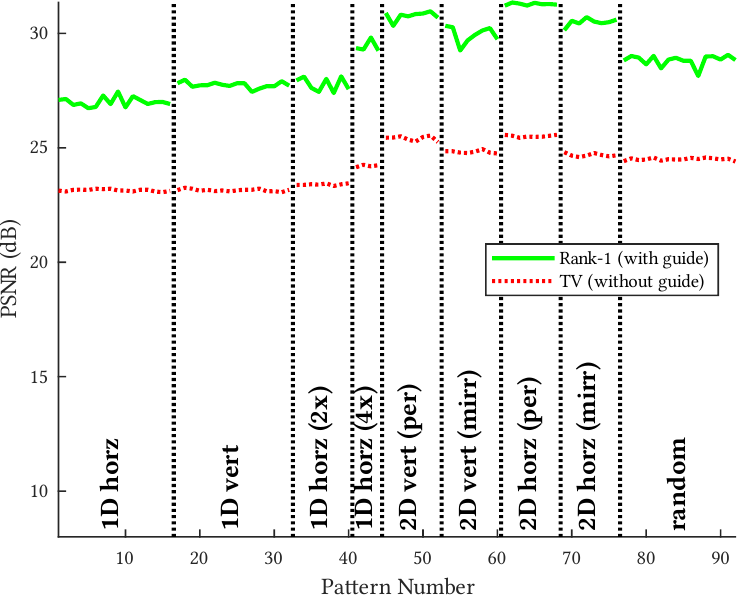}
	\caption{\textbf{Performance of various patterns.} We evaluate rank-1 and TV reconstruction with 92 different patterns, and plot average PSNR for each pattern. Since repeated 2D patterns have a high diversity of spectral profiles in a small neighborhood, the resultant HSI has higher accuracy.}
	\label{fig:sim_pat_idx}
\end{figure}

\textit{Comparisons against snapshot techniques.} We  simulate reconstruction with all 92 patterns to understand their effect on accuracy.
Figure \ref{fig:sim_pat_idx} shows a plot of average PSNR for six test HSIs for various pattern types with rank-1 and TV reconstruction. Since 2D patterns have diverse spectral filters in a local neighborhood, the reconstruction is expected to be more accurate than other patterns, as verified by the plot.

\begin{figure}
	\centering
	\includegraphics[width=\columnwidth]{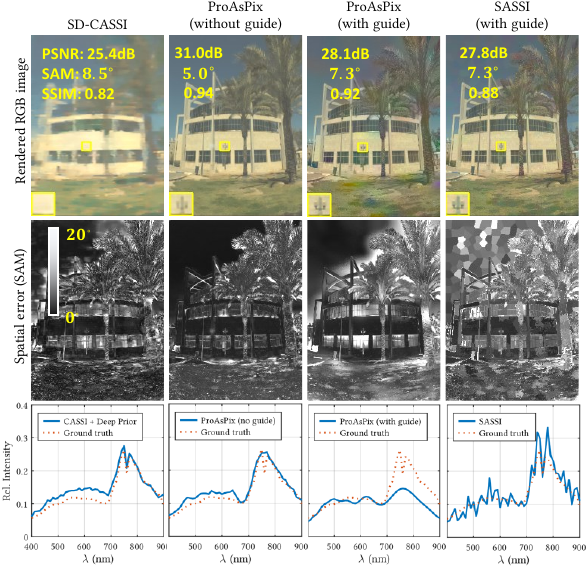}
	\caption{\textbf{Comparisons for broadband imaging.} The range of spectral filters at each pixel of the SLM enables broadband imaging from visible to NIR wavelengths (400 - 900nm). The image above shows (top row) rendered RGB, (middle) spatial error map and (bottom) spectrum at one location for four approaches. Hyperspectral image with SD-CASSI and ProAsPix (without guide) involved a snapshot measurement of spatio-spectral image and recovered with a deep image prior~\cite{ulyanov2018deep}. ProAsPix (with guide) involved a rank-1 recovery. Both ProAsPix and SASSI required an extra RGB image. ProAsPix with and without guide outperform competing, primarily enabled by independent per-pixel spectral coding -- implying that ProAsPix is a superior choice when it comes to broadband imaging.}
	\label{fig:snapshot_visnir}
\end{figure}

\paragraph{Snapshot reconstruction for broadband imaging.}
ProAsPix is an appealing choice for broadband imaging over visible to near IR (NIR) wavelengths.
We were unable to find learned models fine tuned for this range of wavelengths, and hence we compared reconstruction from SD-CASSI~\cite{wagadarikar2008single} measurements, ProAsPix with and without guide images, and SASSI~\cite{saragadam2018krism}.
For SD-CASSI, we employed a DIP-based framework to regularize the linear inverse problem which resulted in a significantly better performance than TV-based recovery~\cite{bioucas2007new}.
Figure~\ref{fig:snapshot_visnir} shows an example reconstruction of a HSI from the ICVL dataset.
We downsampled the 519 bands to 51 bands from 400 - 900 nm wavelength range for all experiments in this figure.
ProAsPix with and without guide both achieve higher performance metrics than competitors.
We note that while reconstruction with a guide is sometimes worse than without guide, it is significantly faster (order of seconds on CPU) than using DIP (several tens of minutes on GPU) making it amenable to real-time reconstruction.
SD-CASSI suffered from smearing of the spatial features due to spatio-spectral mixing.
SASSI suffered from noise in reconstruction as the measurements are sparse in nature.
In contrast, ProAsPix resulted in visually pleasing results as the spectral modulation was done in a dense per-pixel manner.
Techniques such as CASSI and SASSI are aimed solely at hyperspectral imaging. In contrast, ProAsPix enables a larger gamut of applications that go beyond hyperspectral imaging.
In further sections we show two key applications enabled by ProAsPix setup that is not achievable by either CASSI or SASSI setup, namely implementation of an array of spectral filters for optical material classification, and realization of arbitrary spectral filters (similar to tunable filters) at extremely high frame rates.

%

\begin{figure}
\centering
	\includegraphics[width=0.475\columnwidth]{figures/sim/sweep_nmeas_psnr.pdf}
	\includegraphics[width=0.475\columnwidth]{figures/sim/sweep_nmeas_sam.pdf}
	\includegraphics[width=\columnwidth]{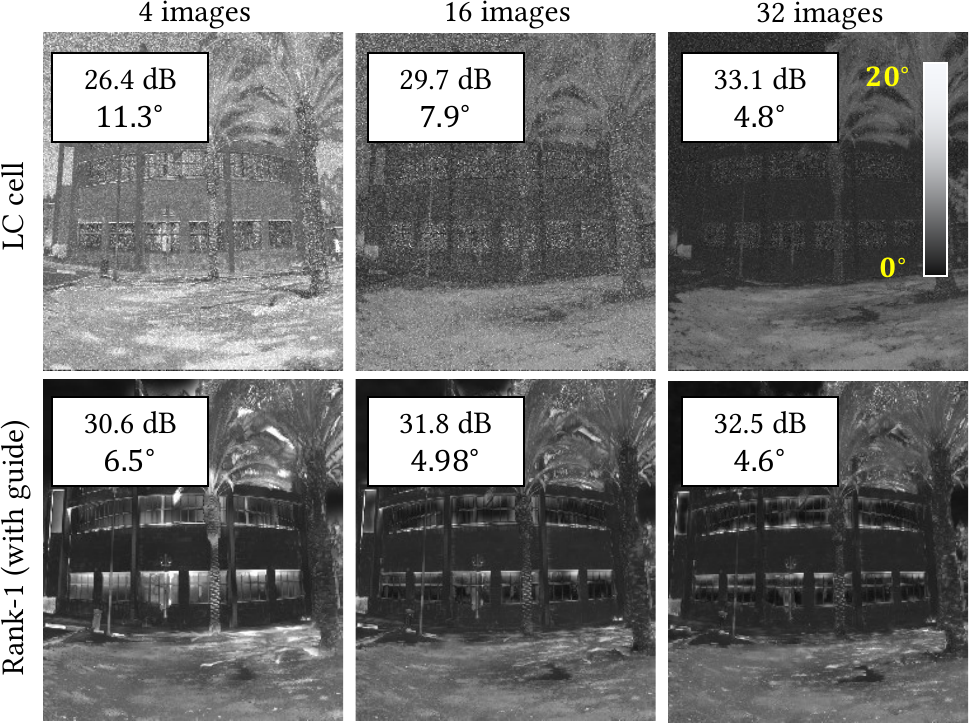}
	\caption{\textbf{Multi-frame performance.} Our proposed approach performs better than CS-MUSI with small number of images or lower light levels. Performance increases uniformly with increasing number of images and light levels, and outperforms CS-MUSI on an average. With 256 images, performance is same as CS-MUSI -- this is to be expected as the measurements with proposed approach and with LC Cell are equivalent.}
\label{fig:sim_multiframe}
\end{figure}

\begin{figure}
	\centering
	\includegraphics[width=0.475\columnwidth]{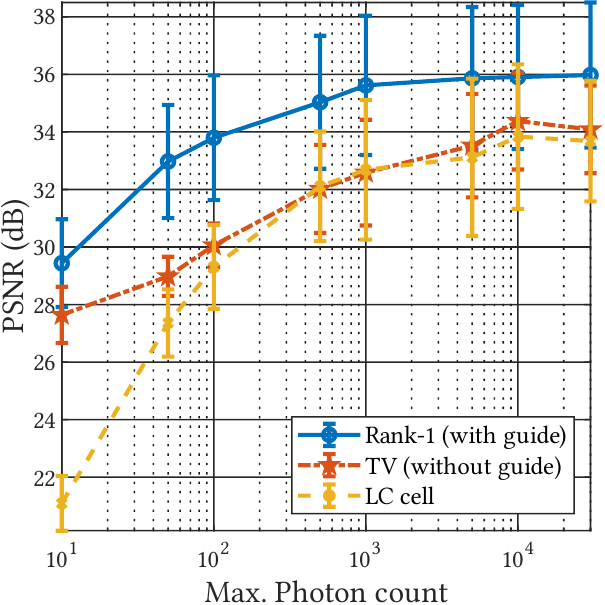}
	\includegraphics[width=0.475\columnwidth]{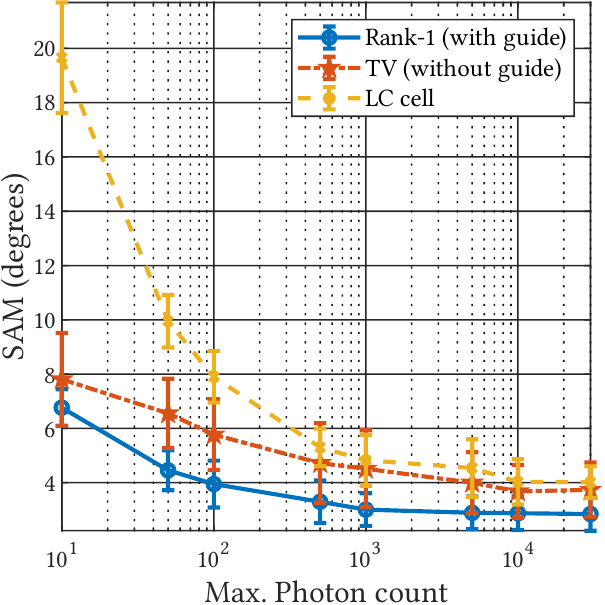}
	\caption{\textbf{Performance with varying light levels.} Experiments were performed with 16 images for all approaches. The plots show PSNR and angular error as a function of light level, while the bottom images show angular error map for LC cell-based reconstruction and our rank-1 reconstruction. At very low to medium light levels, rank-1 outperforms LC cell-based measurements by a large margin. When the light levels are very high, LC cell has a better accuracy, which can be attributed to accurate spatial reconstruction.}
	\label{fig:sim_multiframe_tau}
\end{figure}

\textit{Comparisons with multiple captures.} Our primary competitor for multi frame reconstruction is CS-MUSI~\cite{oiknine2019compressive} where images are captured with a spatially invariant spectral modulation.
Figure \ref{fig:sim_multiframe} plots reconstruction accuracy for CS-MUSI and our technique for varying number of images.
Evidently, our approach with guided filtering produces superior reconstruction than CS-MUSI for all number of measurements.
We do observe that the reconstruction in the absence of a guide image is better than CS-MUSI at fewer measurements and similar with 50 or more images.
For a small number of images, it is more advantageous to spatially multiplex the various spectral filters.
However with a large number of images, spatial multiplexing has a similar effect to capturing images with spatially invariant spectral filters.
We also compared CS-MUSI against our technique with varying light levels in Fig.~\ref{fig:sim_multiframe_tau}.
We noticed that our approach is more robust to noise, especially at very low to low light levels -- this advantage primarily arises from the use of guide image which acts as a regularizer for the spatial dimension.

\subsection{Experiments with lab prototype}

\begin{figure}
	\centering
	\includegraphics[width=0.225\textwidth,trim=4.75in 0 5.6in 0.625in, clip]{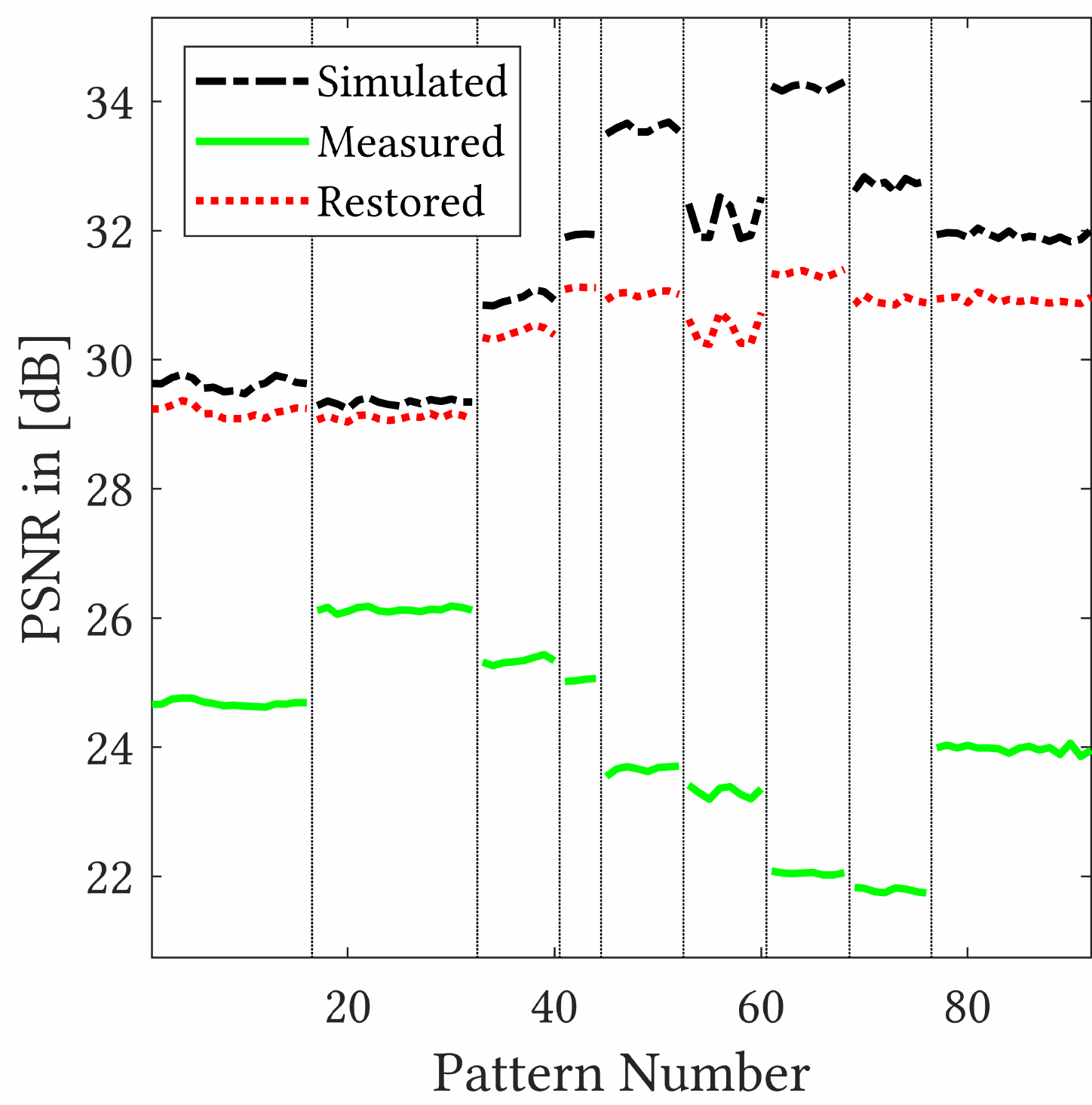}
	\includegraphics[width=0.225\textwidth,trim=4.75in 0 5.6in 0.625in, clip]{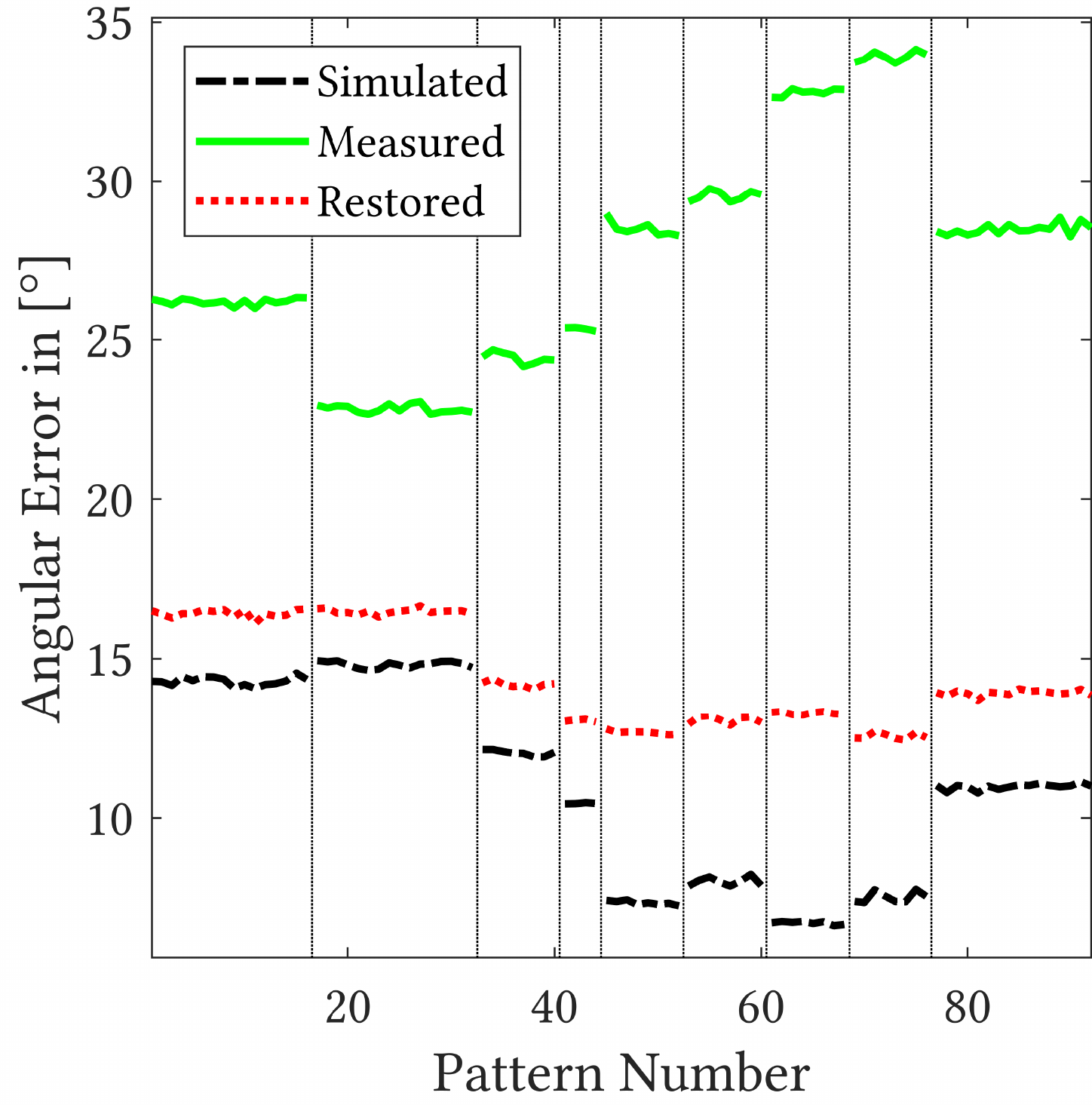}
	\caption{\textbf{Performance on real data over different patterns.} We evaluate the performance of rank-1 filtering on the 92 different patterns over the test scenes. Shown are the average PSNR and angular error for each of the 92 patterns. Generally, 2D patterns outperform 1D patterns, due to the richer set of local filters in such patterns. The results with raw measurements are similar in spirit to prior work by \citet{Zhu:13}. Accounting for the optical aberrations dramatically improves the performances, often times by 10 dB.}
	\label{fig:patternresults2}
\end{figure}

\paragraph{HSI reconstruction from a single spectrally-coded image.} We first evaluate the reconstruction from a single spectrally-coded image. 
The performance of TV-based reconstructions were significantly worse than rank-1 reconstructions, where the auxiliary RGB guide image is used. 
The results obtained with deep image prior were better than TV, but required considerable amount of GPU memory, precluding reconstruction of our real results.
So we limit the the results to just the rank-1 reconstruction here.
Figure \ref{fig:patternresults2} provides the reconstruction performance of the guided rank-1 reconstruction for each of the 92 patterns in Figure \ref{fig:restore_patterns}.
We characterize performance with PSNR, measured in dB, and spectral angular error, measured in degrees.
The spectral angular error is computed as the median angular mismatch between the estimated and ground truth spectrum observed at a pixel, as defined in (\ref{eq:sam}).
For the ground truth, we use the full scan data, acquired under 256 constant-valued SLM patterns; we use a least squares solver to recover the spectrum at each pixel individually.
All plots provide aggregate results over the test scenes  in Figure \ref{fig:dataset}.

\begin{figure}
	\centering
	\includegraphics[width=0.475\textwidth]{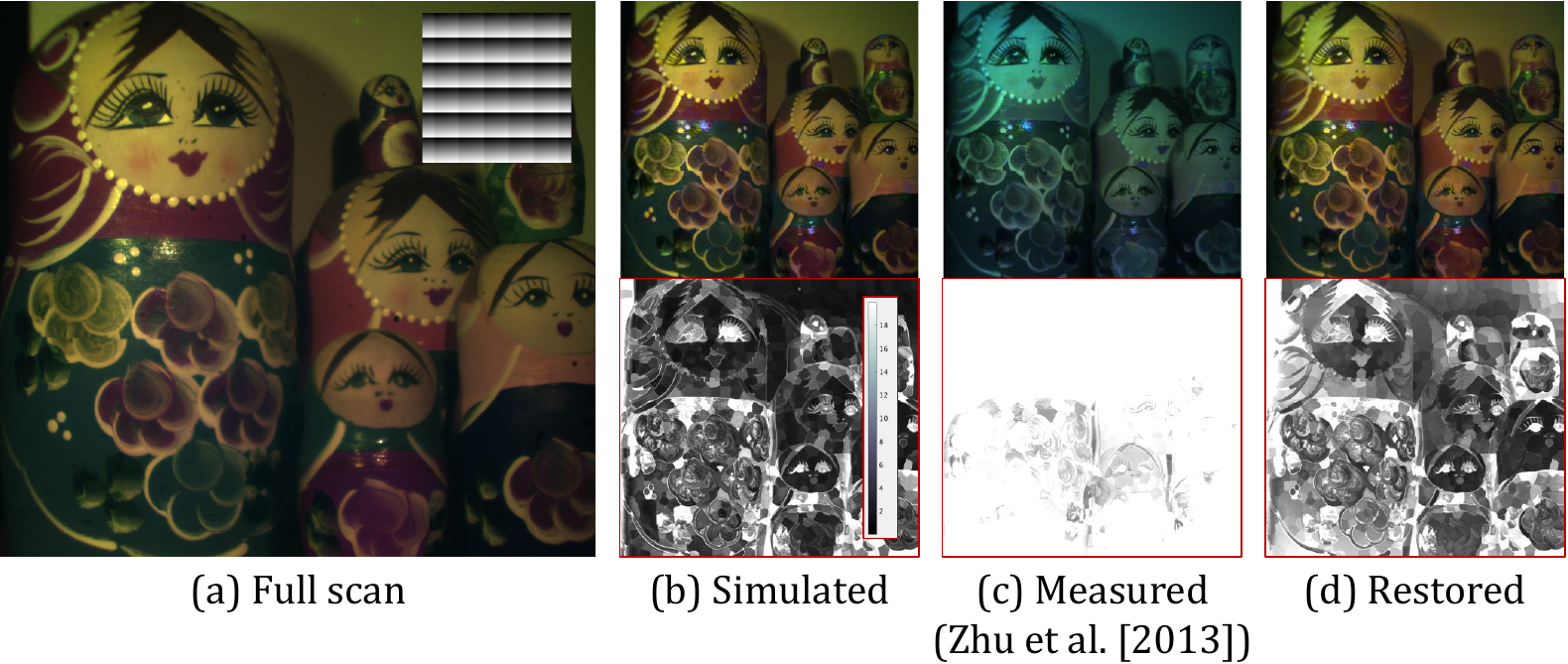}
	\caption{\textbf{Comparison of simulated, measured, and restored measurements.} (a) Shown is an RGB image of the scene with inset of the spectral filter array that we used for the results in the other columns. (b-d) We visualize reconstructed HSIs using the simulated, measured (\cite{Zhu:13}), and restored measurements as  rendered RGB images. Below each RGB reconstruction, we provide the angular error against the full scan reconstructions; for these error maps, the brightest values are errors that are $20^\circ$ or higher. }
	\label{fig:patternresults3}
\end{figure}

For each of the 92 patterns, we evaluate the performance using the raw measurements from the camera, the simulated data where we use the full scan data to mimic the measurement process, and the restored measurements use the deep neural network described in Section \ref{sec:restorenet}.
The restored measurements generally outperform the raw measurements, often by a very large margin; this is fully consistent with our earlier observation in Figure \ref{fig:restore_patterns}, that the restoration network do reduce the mismatch to ground truth. 

\begin{figure}
	\centering
	\includegraphics[width=0.475\textwidth]{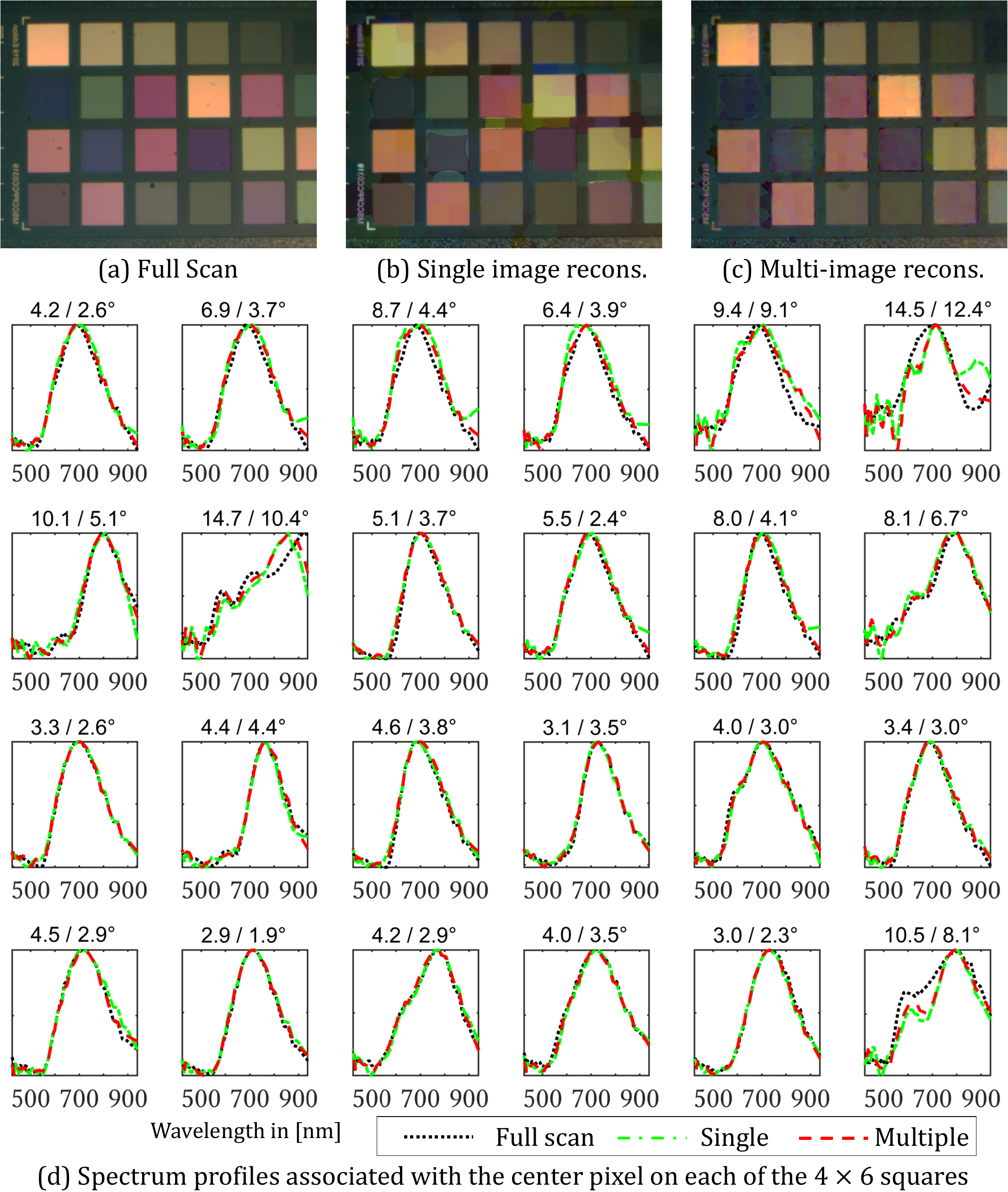}
	\caption{\textbf{Color checkerboard under an incandescent lamp.} (a, b, c) Rendered RGB image for the full scan and restored measurements with single and 16 measurements. For the single image. we used the 2D horizontal/periodic pattern as the SLM pattern. (d) For each of the $4 \times 6$ squares, we plot the reconstructed SNR at the center of each square. The number on top of each plot indicates the spectral angular error between the estimated and ground truth spectra. The first number is for single image and the second for the 16 image reconstruction. We can observe a very high level of match to the ground truth. Given that the first row has the same spectrum under different light levels, we can get a sense of the noise performance of our approach (and of the full scan reconstructions).}
	\label{fig:checker}
\end{figure}

\paragraph{Color checkerboard.} Figure \ref{fig:checker} visualizes the performance of the proposed technique on a color checkerboard illuminated with an incandescent lamp. 
We show reconstructions using the rank-1 guided filter for a single measurement, with the first pattern in Figure \ref{fig:whatpatterns}, as well as with 16 measurements.
We compare the spectrum performance against the full scan reconstruction and observe a high level of match in the spectrum between the methods, with multi-image reconstruction performing consistently better.

\begin{figure}
	\centering
	\includegraphics[width=0.235\textwidth]{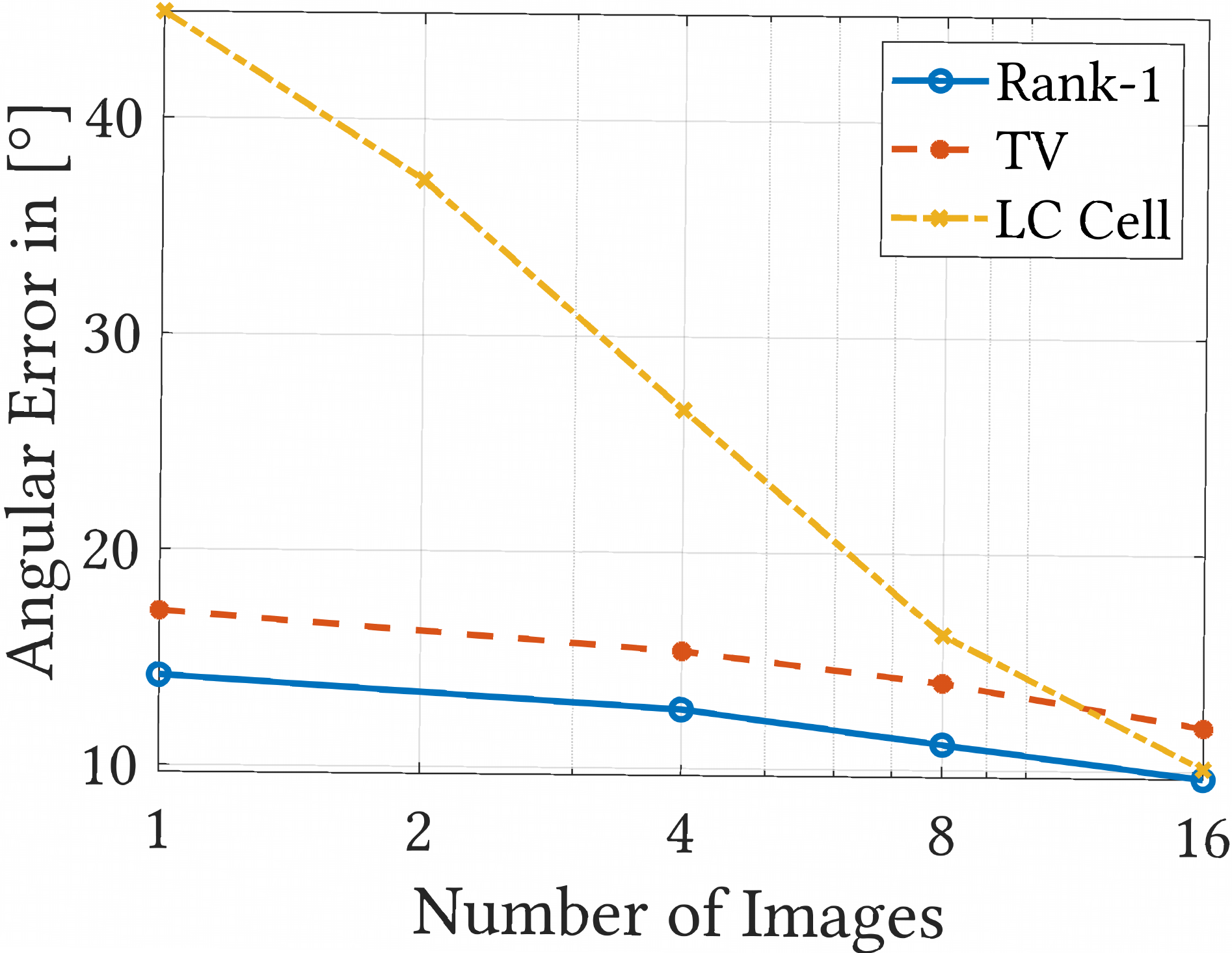}
	\includegraphics[width=0.235\textwidth]{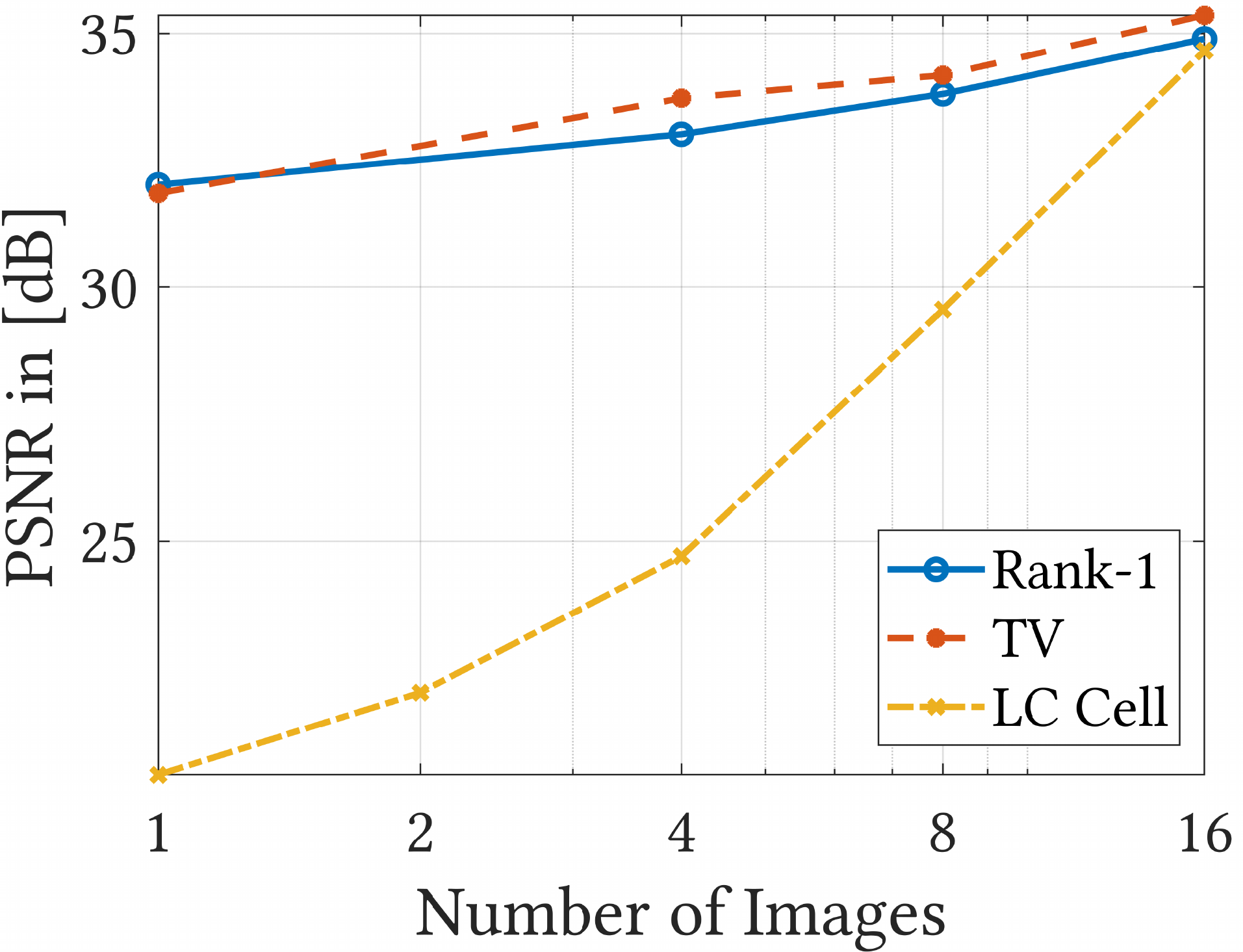}
	\caption{\textbf{Performance with multiple patterns.} We provide angular error and PSNR plots with increasing number of measurements for different measurement and reconstruction techniques. For the proposed method, we evaluate the performance of the rank-1 guided filter, which uses the auxiliary guide RGB, and the guide free TV-prior. We can observe that the performance of TV approaches that of the guided filter as the number of images increases. We also compare to a measurement strategy where we turn the SLM into an LC cell, by displaying constant intensity pattern on it. For a number of measurements smaller than 8, the paucity of spectral filters leads to poorer performance when compared against techniques that benefit from having spatially-varying patterns on the SLM. The numbers for the LC cell are average values over 10 different random selections of grayscale intensities. }
	\label{fig:multiplot}
\end{figure}

\paragraph{Color checkerboard.} Figure \ref{fig:checker} visualizes the performance of the proposed technique on a color checkerboard illuminated with an incandescent lamp. 
We show reconstructions using the rank-1 guided filter for a single measurement, with the first pattern in Figure \ref{fig:whatpatterns}, as well as with 16 measurements.
We compare the spectrum performance against the full scan reconstruction and observe a high level of match in the spectrum between the methods, with multi-image reconstruction performing consistently better.

\subsection{Discussions}
\begin{figure}[!tt]
	\centering
	\includegraphics[width=0.4\textwidth]{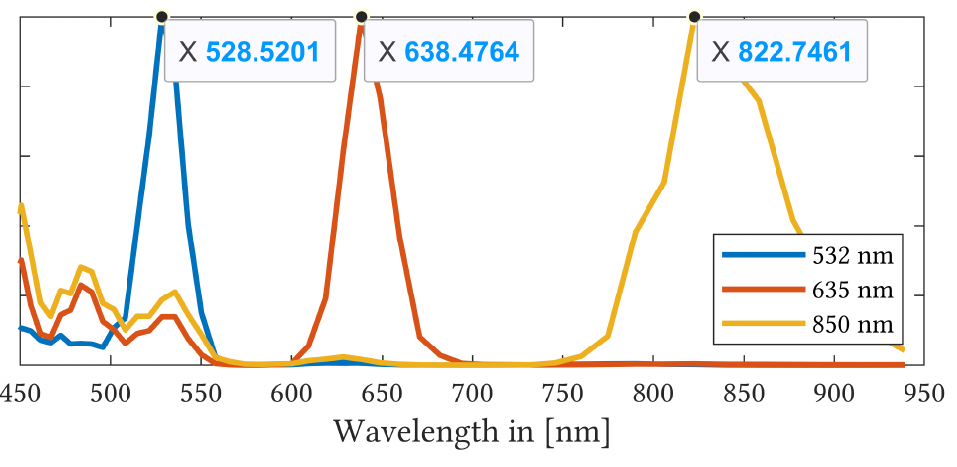}
	\caption{\textbf{Spectral resolution.} To characterize the spectral resolution, we visualize the reconstructions, with full scan measurements, of three lasers. We obtain a FWHM  bandwidths of  $23.9 $nm at $532$nm, $31.9$nm at $635$nm and $69.0$nm at $850$nm. As is consistent with LC cells, the device obtains a  bandwidth that is roughly proportional to the wavelength.}
	\label{fig:laser}
\end{figure}


\paragraph{Spectral resolution.} The spectral resolution of our prototype is largely similar to what we expect with an LC cell-based device; specifically, we get a nonlinear spectral resolution with high resolution at smaller wavelength.
We verify this with a scene illuminated with three lasers, and provide the reconstructions in Figure \ref{fig:laser}.

\begin{figure}[!tt]
	\centering
	\includegraphics[width=0.4\textwidth]{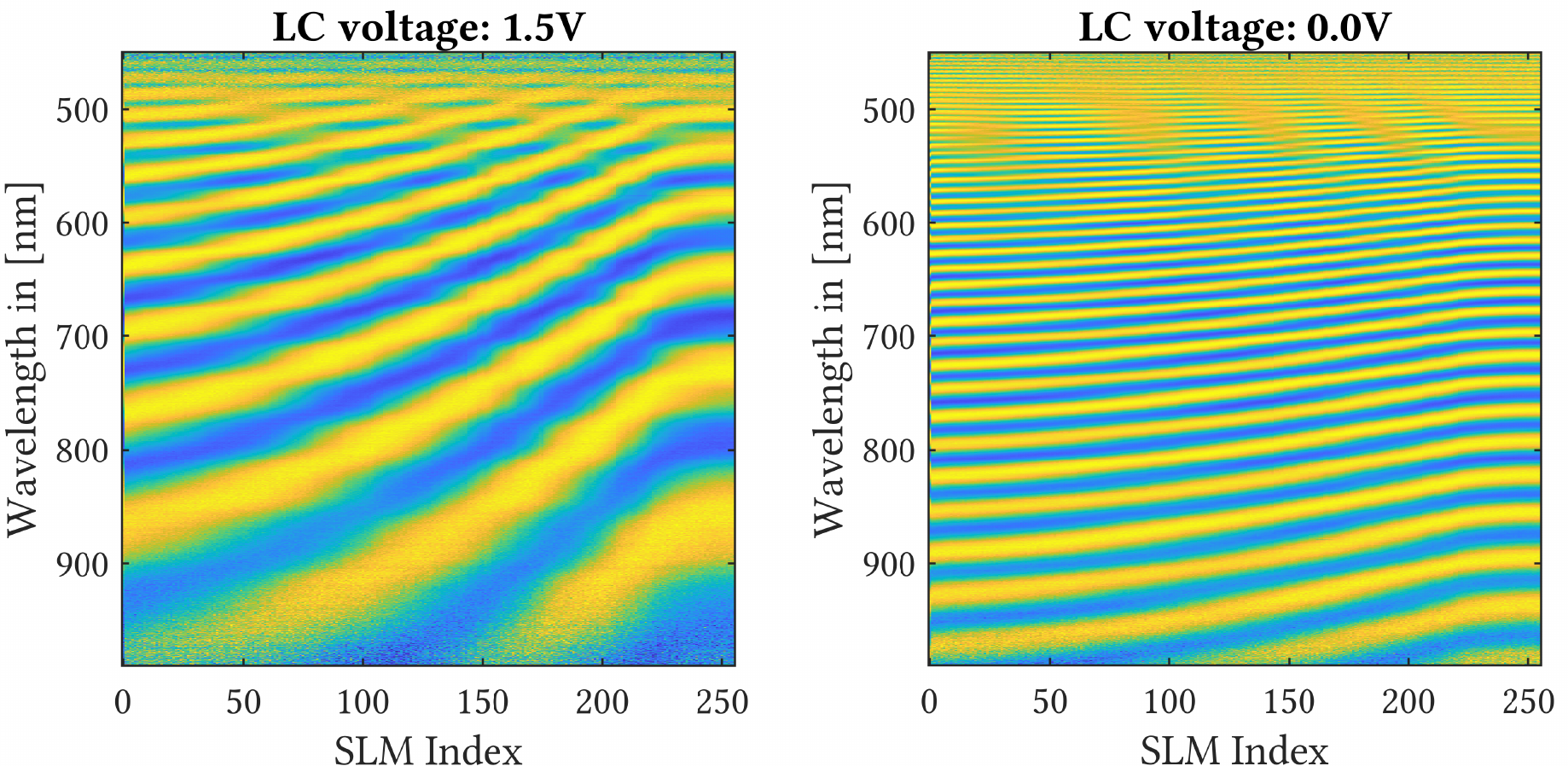}
	\caption{\textbf{Increasing range of spectral modulation using an LC cell in the optical pathway.} We can use an additional LC cell to increase the overall phase retardance, and hence the diversity of spectral filters implemented by our system. Shown are the filters implemented by the SLM at two different input voltages across the LC cell.}
	\label{fig:LC}
\end{figure}

\paragraph{Enhancing spectral diversity of filters.} The richness of the spectral modulation produced by our system relies critically on the range of phase retardation that can be implemented by the SLM.
For our system, this range spans $3 \mu m$ to $800 nm$ --- increasing this range is an important direction in enhancing the utility of our design.
One way of realizing this is by introducing an LC cell in front of the SLM and using the additional phase retardation provided by it.
Figure \ref{fig:LC} visualizes the range of spectral filters we can obtain once we add such an LC retarder (Thorlab LCC1115-B) immediately in front of the SLM. 
The resulting setup implements spectral filters that  have the form
\[ \frac{1}{2} - \frac{1}{2} \cos\left( 2\pi \frac{\Delta n(v_{\textrm{SLM}}(x, y)) d_{\textrm{SLM}} + 2 n(v_{\textrm{LC}}) d_{\textrm{LC}} }{\lambda} \right), \]
where the terms marked with ``SLM'' and ``LC'' denote to retardances applied at the SLM and LC cell, respectively.
Controlling the voltage across this retarder shifts the range of phase retardance achievable with our setup.
Since the LC cell has the largest birefringence when no voltage is applied across it, we see the spectral filters with the largest oscillations at low voltages; however, the diversity of filters is low in this setting since the added phase retardance overwhelms the range of the SLM.
At a slightly higher voltage (1.5V in Figure \ref{fig:LC}), we observe a different set of filters with greater diversity but fewer cycles.
Hence, adding an LC cell opens up a novel and richer design space  and is likely a powerful addendum to our design.

%

\end{appendices}

{\small
\bibliographystyle{plainnat}
\bibliography{refs}
}

\end{document}